%% file: spot.tex
\documentclass[runningheads]{llncs}
\usepackage{graphicx}

\usepackage{hyperref}
\usepackage{tikz}
\usepackage{comment}
\usepackage{amsmath,amssymb} 
\usepackage{color}
\usepackage{booktabs}
\usepackage{epsfig}
\usepackage{amsmath}
\usepackage{amssymb}
\usepackage{xspace}
\usepackage{xcolor}
\usepackage{tabularx}
\usepackage{colortbl}
\usepackage{multirow}
\usepackage{rotating}
\usepackage{mathtools}
\usepackage{placeins}
\usepackage{paralist}
\usepackage{subcaption}
\usepackage{cite}

\usepackage[accsupp]{axessibility}  

\usepackage[width=122mm,left=12mm,paperwidth=146mm,height=193mm,top=12mm,paperheight=217mm]{geometry}

\input{src/notation.tex}

\begin{document}
\pagestyle{headings}
\mainmatter

\title{Spotting Temporally Precise, Fine-Grained Events in Video}

\author{
    James Hong\inst{1} \and
    Haotian Zhang\inst{1} \and
    Micha\"el Gharbi\inst{2} \and \\
    Matthew Fisher\inst{2} \and
    Kayvon Fatahalian\inst{1}
}
\authorrunning{J. Hong et al.}
%
\institute{Stanford University \and Adobe Research}

\maketitle

\input{src/abstract.tex}

\input{src/intro.tex}

\input{src/related.tex}

\input{src/method.tex}

\input{src/dataset.tex}

\input{src/result.tex}

\input{src/discussion.tex}

\input{src/conclusion.tex}

\input{src/ack.tex}

\appendix

\input{src/supp_impl.tex}

\input{src/supp_result.tex}

\input{src/supp_dataset.tex}

\clearpage
%
%
\bibliographystyle{splncs04}
\bibliography{spot}
\end{document}

%% file: src/notation.tex
\newcommand{\notation}[1]{\ensuremath{#1}\xspace}

\newcommand{\OURMETHOD}{{E2E-Spot}\xspace}

\newcommand{\fscomp}{{FS-Comp}\xspace}
\newcommand{\fsperf}{{FS-Perf}\xspace}
\newcommand{\tennis}{{Tennis}\xspace}
\newcommand{\finegym}{{FineGym}\xspace}
\newcommand{\finediving}{{FineDiving}\xspace}
\newcommand{\soccernet}{{SoccerNet-v2}\xspace}

\newcommand{\NumFrames}{\notation{N}}
\newcommand{\NumClasses}{\notation{K}}
\newcommand{\Tolerance}{\notation{\delta}}

\newcommand{\Time}{\notation{t}}
\newcommand{\Frame}[1]{\notation{\mathbf{x}_{#1}}}
\newcommand{\Class}[1]{\notation{\mathbf{c}_{#1}}}
\newcommand{\Label}[1]{\notation{\mathbf{y}_{#1}}}
\newcommand{\Prediction}[1]{\notation{\hat{\mathbf{y}}_{#1}}}

\newcommand{\AllFrames}{\notation{\Frame{1},\ldots,\Frame{\NumFrames}}}
\newcommand{\AllClasses}{\notation{\Class{1},\ldots,\Class{\NumClasses}}}
\newcommand{\AllPredictions}{\notation{\Prediction{1},\ldots,\Prediction{\NumFrames}}}

\DeclareMathOperator{\CE}{CE}

\newcommand{\FeatureExtractor}{\notation{F}}
\newcommand{\TemporalArchitecture}{\notation{G}}

\newcommand{\textsuperddagger}{\textsuperscript{\textdaggerdbl}}

\newcommand{\best}[1]{\underline{#1}}
\newcommand{\sota}[1]{\textbf{#1}}

\newcommand{\nms}{\textsuperscript{\textdagger}}

\newcommand{\loss}{*}

%% file: src/abstract.tex
\begin{abstract}

We introduce the task of spotting temporally precise, fine-grained events in video (detecting the precise moment in time events occur).
Precise spotting requires models to reason globally about the full-time scale of actions and locally to identify subtle frame-to-frame appearance and motion differences that identify events during these actions.
Surprisingly, we find that top performing solutions to prior video understanding tasks such as action detection and segmentation do not simultaneously meet both requirements.
In response, we propose E2E-Spot, a compact, end-to-end model that performs well on the precise spotting task and can be trained quickly on a single GPU.
We demonstrate that E2E-Spot significantly outperforms recent baselines adapted from the video action detection, segmentation, and spotting literature to the precise spotting task.
Finally, we contribute new annotations and splits to several fine-grained sports action datasets to make these datasets suitable for future work on precise spotting.

\keywords{temporally precise spotting; video understanding}
\end{abstract}

%% file: src/intro.tex
\input{src/figure/teaser.tex}

\section{Introduction}

Detecting the precise moment in time events occur in a video (temporally precise
event `spotting') is an important video analysis task that stands to be essential to many future advanced video analytics and video editing~\cite{vid2player} applications.
However, despite significant progress in fine-grained video
understanding~\cite{epickitchens,kinetics,finegym,ssv2}, temporal action
detection (TAD)~\cite{activitynet,toyotasmarthome,thumos14,charades,multithumos}, and temporal action segmentation (TAS)~\cite{gtea,breakfast,50salads}, precise event spotting has rarely been studied by the video understanding community.

We address this gap by focusing on the challenge of precisely spotting events in sports video.
We study sports video because of the quantity
of data available and the high temporal accuracy needed to analyze human performances.
For example, we wish to determine the frame in which a tennis player hits
the ball, the frame a ball bounces on the court, or the moment a figure skater starts
or lands a jump.
\autoref{fig:teaser} shows examples from these sports and illustrates why precise spotting is challenging.
The goal is to identify the precise frame when an event occurs, but adjacent frames are extremely similar visually;
looking at one or two frames alone, it can be difficult even for a human to judge when a racket makes contact with a ball or when a figure skater lands a jump.
However, inspection of longer sequences of frames makes the
task significantly more tractable since the observer knows when to expect the
event of interest in the context of a longer action (e.g., the swing of the racket, the preparation for a jump, or a ball's trajectory).
Therefore, we hypothesize that precise spotting requires models that can (1) represent subtle appearance and motion clues, and also (2) make decisions using information spread over long temporal contexts.

Surprisingly, we have found that the large body of literature on video
understanding lacks solutions that meet these two requirements in the regime of temporally precise spotting.
For example, action recognition (classification) models are not designed to operate efficiently on large temporal windows and struggle to learn in the heavily class-imbalanced setting created by precise spotting of rare events.
Sequence models from segmentation and detection extract patterns over longer timescales, but training these complex models end-to-end has led to optimization challenges.
This has resulted in many solutions that operate in two phases, relying on pre-trained (or modestly fine-tuned) input features that are not particularly specialized to capture the subtle (and often highly domain-specific) visual details needed to spot events with temporal precision.

%
We propose a simpler alternative (\OURMETHOD) to satisfy our hypothesized requirements.
The key to training a sequence model end-to-end over a wide temporal context is
an efficient per-frame feature extractor that can process hundreds of contiguous
frames without exceeding platform memory.
We demonstrate how to combine existing modules from the video processing literature to accomplish this goal without introducing new, bespoke architectures.

Despite its simplicity, \OURMETHOD significantly outperforms prior baselines, which opt for a two-phase approach, as well as naive end-to-end learning approaches on precise spotting.
Moreover, \OURMETHOD is computationally efficient at inference time
and can complete the full end-to-end spotting task in less time than just the
feature extraction phase of many prior methods~\cite{tsp,i3d}.

This paper makes three main contributions:
\begin{enumerate}
    \item The novel task of temporally precise spotting of fine-grained events.
    We introduce frame-accurate labels for two existing fine-grained sports action datasets: Tennis~\cite{vid2player} and Figure Skating~\cite{vpd}.
    We also adapt the temporal annotations from FineGym~\cite{finegym} and FineDiving~\cite{finediving} to show the generality of the precise spotting task.

    \item \OURMETHOD, a from-the-ground-up, end-to-end learning approach to
    precise spotting that combines well-established architectural
    components~\cite{gatedrnn,regnet,gsm} and can be trained quickly on a
    single GPU.

    \item Analysis of spotting performance.
    \OURMETHOD outperforms strong baselines (\autoref{sec:result}) on precise temporal spotting (by 4--11 mAP, spotting within 1 frame).
    \OURMETHOD is also competitive on coarser spotting tasks (within 1--5 sec), achieving second place in the 2022 SoccerNet Action Spotting challenge~\cite{soccernetv2,snspotting} (within 1.1 avg-mAP) and a lift of 14.8--16.5 avg-mAP over prior work.
\end{enumerate}

Our code and data are publicly available.

%% file: src/figure/teaser.tex
\begin{figure}[t]
	\centering
	\includegraphics[width=\columnwidth]{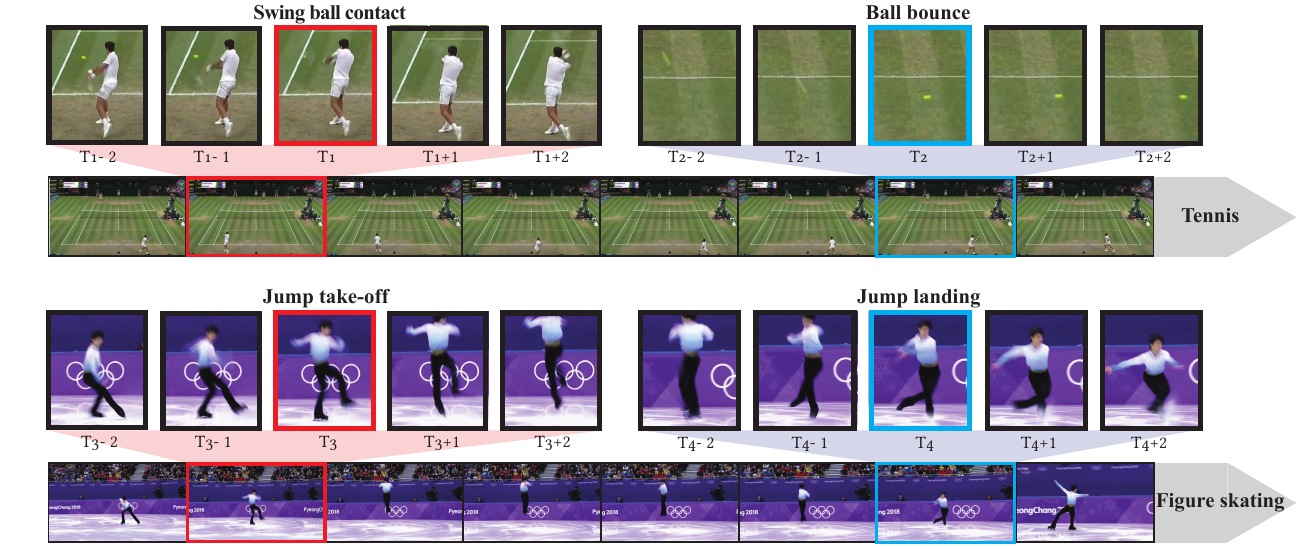}
	\caption{We perform temporally precise spotting of events in video, where success requires detecting the occurrence of an event within a single or small tolerance of frames.
	Examples of precise events: in \textit{tennis}, the moment a player contacts the ball during a swing (red) or when a ball bounces on the court (blue); in \textit{figure skating}, the moment of take-off (red) and landing (blue) during a jump.
	}
	\label{fig:teaser}
\end{figure}

%% file: src/related.tex
\section{Related Work}
\label{sec:related}

\subsubsection*{Action Spotting.}
Previous work on spotting~\cite{soccernetv2} focuses
on \emph{coarse} action spotting, where a detection is deemed correct if it
occurs within some time-window around the true event, with a loose error
tolerance (1--5 or 5--60 seconds, equating to 10--100s of frames).
On the Tennis~\cite{vid2player} and Figure Skating~\cite{vpd} datasets described in~\autoref{sec:dataset}, a spotting
error larger than 1--2 frames is essentially equivalent to missing the event
altogether (e.g., a ball impact's on the ground;~\autoref{fig:teaser}).
For demanding applications that require precise temporal annotations,
we argue the relevant task is \emph{precise} event spotting, where
detection thresholds are much more stringent tolerances (1--5 frames; as little as 33 ms in 25--30 FPS video).
We use a similar metric to coarse action spotting: mean Average Precision (mAP @ $\Tolerance$) but with a short temporal tolerance \Tolerance.

\subsubsection*{Temporal Action Detection (TAD) and Segmentation (TAS)} localize \emph{intervals}, often spanning several seconds and containing an `action'. Depending on the dataset, these can be atomic actions such as ``standing up''~\cite{charades} or broad activities such as ``billiards''~\cite{thumos14}.
For such action definitions, it is often unclear what would be considered a temporally precise event to spot.

The success criteria for TAD and TAS also differ from that of precise spotting.
TAD~\cite{activitynet,toyotasmarthome,thumos14,charades,multithumos} is evaluated on interval-based metrics such as mAP @ temporal Intersection-over-Union (IoU) or at sub-sampled time points, neither of which enforce frame accuracy on the action boundaries.
Down-sampling in time (up to 16$\times$) is a common preprocessing step~\cite{sstad,bmn,bsn,charadeschallenge,gtad,actionformer}.
TAS~\cite{gtea,breakfast,50salads} also optimizes interval-based metrics such as F1 @ temporal overlap.
Frame-level metrics for TAS reward accuracy on densely labeled, intra-segment frames; in contrast, event frames in our spotting datasets are sparse.
Spatial-temporal detection benchmarks~\cite{avakinetics,multisports} differ from standard TAD, TAS, and precise spotting by combining both spatial and temporal IoU~\cite{multisports}.

Recent approaches for TAD~\cite{pdan,bmn,bsn,mlad,gtad,pgcn} and TAS~\cite{hasr,sstda,mstcn,asrf,c2ftcn,asformer} often proceed in two stages: (1) feature extraction then (2) head learning for the end task.
Fixed, pre-trained features from video classification on Kinetics-400 are often used for the first stage~\cite{tsp,i3d,tsn}, and
state-of-the-art TAD methods with these features~\cite{muses,actionformer,vsgn} often perform comparably to if not better than recent end-to-end learning approaches~\cite{afsd,e2etad}.
Indirect fine-tuning using classification in the target domain is sometimes performed to improve feature encoding~\cite{tsp,charadeschallenge}.
Early end-to-end approaches encode video as non-overlapping segments~\cite{sstad} (e.g., 16 frames) or downsample in time~\cite{asynctemporalfields,multistreamfgdet}, producing features that are too temporally coarse to be effective for spotting frame-accurate events.

Like TAD and TAS, precise spotting is a temporal localization task performed on untrimmed video.
As is the case, many models for TAD and TAS can be adapted for precise spotting.
We use MS-TCN~\cite{mstcn}, GCN~\cite{gtad}, GRU~\cite{gatedrnn}, and ASFormer~\cite{asformer} as baselines, and we test these models with different  features~\cite{tsp,i3d,mvit} in~\autoref{sec:result}.
However, we find that relying on fixed or indirectly fine-tuned features as input for these models is a critical limitation.
Our experiments show that (1) \OURMETHOD is a strong baseline for precise spotting and (2) more complex architectures do not necessarily provide additional benefit when feature learning is end-to-end.
Finally, we note the long history of CNN-RNN architectures in TAD/TAS~\cite{sstad,sst,msbilstm,multithumos,tricornet};
\OURMETHOD is a simple design from this family, motivated by our requirements for frame-dense processing and end-to-end learning, and implemented using a modern CNN for spatial-temporal feature encoding.

\subsubsection*{Video Classification} predicts one label for an entire video, as opposed to per-frame labels for spotting.
This leads to two key differences: (1) sparsely sampling frames~\cite{slowfast,tsn} is effective, whereas precise spotting requires dense sampling;
(2) to obtain a video-level prediction, popular architectures for classification typically perform global space-time
pooling~\cite{r21d} or temporal consensus~\cite{tsm,tsn,trn}.
\OURMETHOD shows that omitting temporal pooling\footnote{Omission of temporal pooling is similar to concurrent work, E2E-TAD~\cite{e2etad}.} and training end-to-end yields an efficient pipeline for precise, per-frame spotting.

\OURMETHOD incorporates ideas from popular video classification models for spatial-temporal feature extraction.
TSM~\cite{tsm} introduced the temporal shift operation, which converts a 2D CNN into a spatial-temporal feature extractor by mixing channels between time steps.
GSM~\cite{gsm} learns the shift.
We find the combination of RegNet-Y~\cite{regnet} and GSM~\cite{gsm} to be effective and suggest these building blocks as a starting point for future spotting research.

\subsubsection*{Sports Activity Datasets} are a fertile testing ground for
video action recognition and understanding~\cite{soccernetv2,cbdqsa,vpd,volleyball,diving48,multisports,finegym,finediving,vid2player}.
We evaluate using temporal annotations
from several recent datasets~\cite{soccernetv2,vpd,finegym,finediving,vid2player}.
These datasets are fine-grained, meaning that all event and class labels relate to a single activity (i.e., a single sport), as compared to coarse-grained datasets~\cite{activitynet,thumos14}, where classes comprise a broad mix of generic activities.
Supporting fine-grained concepts and labels is an important requirement of many practical, real-world applications.

%% file: src/method.tex
\input{src/figure/pipeline.tex}

\section{\OURMETHOD: An End-to-End Model for Precise Spotting}
\label{sec:method}

We define the precise temporal event spotting task as follows: given a
video with $\NumFrames$ frames $\AllFrames$ and a set of $\NumClasses$ event
classes $\AllClasses$, the goal is to predict the (sparse) set of frame indices when an event occurs, as well as the event's class $(\Time,\Prediction{t})\in\mathbb{N} \times\{\AllClasses\}$.
A prediction is deemed correct if its timestamp falls within
\Tolerance frames of a labeled ground-truth event and it has the correct class label.
In precise spotting, the temporal tolerance \Tolerance is small --- i.e., a few
frames only.
We assume that the frame rate of the video is sufficiently high to capture the precise event and that frame rates are similar across videos.

We identified several key design requirements for a model to
perform well on the temporally precise spotting task:
\begin{enumerate}
    \item Task-specific {\bf local spatial-temporal features} that capture subtle visual differences and motion across neighboring frames.

    \item A {\bf long-term temporal reasoning} mechanism, which
    provides a long temporal window to spot short, rare events.
    For instance, it is difficult to identify the precise time a figure skater
    enters a jump from a handful of frames.
    But spotting becomes much less ambiguous given the wider context of the acceleration (before) and landing (after the jump) (see~\autoref{fig:teaser}).
    These contexts can occur over many seconds and frames.

    \item {\bf Dense frame prediction} at the temporal granularity of a single frame.

\end{enumerate}
These requirements call for an expressive and efficient network architecture that can be trained end-to-end via direct supervision on spotting.

\OURMETHOD treats a video classification network (with global temporal pooling removed) as part of a sequence model, so that processing a clip of \NumFrames frames results in \NumFrames output features and \NumFrames per-frame predictions.
\autoref{fig:pipeline} illustrates our pipeline.
Frames from each RGB video are first fed to a local spatial-temporal
feature extractor \FeatureExtractor, which produces a dense feature vector
for each frame (\autoref{sub:st_features}).
This lightweight feature extractor incorporates Gate Shift Modules (GSM)~\cite{gsm} into a
generic 2D convolutional neural network (CNN)~\cite{regnet}.
The feature sequence is then further processed by a sequence model
\TemporalArchitecture, which builds a long-scale temporal context
and outputs a class prediction for every frame, including a `background' class to indicate when no event was detected (\autoref{sub:longterm_reasoning}).

\subsection{Local Spatial-Temporal Feature Extractor, $\FeatureExtractor$}
\label{sub:st_features}
The first stage of our pipeline extracts spatial-temporal features for each frame.
We strive to keep the feature extractor as lightweight as possible, but found
that a simple 2D CNN that processes frames
independently~\cite{calf,netvladpp,rmsnet,tsn} is often insufficient for precise
spotting (see~\autoref{sub:ablation}).
This is because a 2D CNN does not capture the spatially-local temporal correlations between frames.
In videos that are densely sampled (24--30 FPS), this temporal signal is
critical to learn features that can robustly differentiate otherwise very
similar frames:
for instance, the speed and travel direction of a tennis ball, when each frame
likely exhibits motion blur.
To obtain more expressive, motion-sensitive features we implement
$\FeatureExtractor$ as a 2D CNN with Gate Shift Modules (GSM)~\cite{gsm}.
We choose RegNet-Y~\cite{regnet}, a recent and compact CNN, as the 2D backbone.

Our feature extractor is similar to models for
video classification~\cite{tsm,gsm,tsn}, but with two key differences:
(1) it samples frames \emph{densely} and
(2) it uses no final temporal consensus/pooling because our goal is to obtain one output per frame, rather than one for the whole video or multi-frame segment.

\subsubsection*{Efficiency Compared to Other Per-frame Feature Extractors.}
A common alternative for per-frame feature extraction~\cite{tsp,mstcn} is to stride a video classification model densely --- i.e., by using a model which takes $M$ frames as input and produces a single feature and by running it on the $M$ frame neighborhood of every frame.
The overhead of processing each frame multiple
times in overlapping windows makes end-to-end feature learning or fine-tuning difficult for tasks like spotting that require dense processing of frames.
In contrast, our approach processes each frame once and can be trained as part of an end-to-end pipeline with much longer sequences (100s of frames), even on a single GPU (see~\autoref{tab:model_throughput}).

\subsection{Long-term Temporal Reasoning Module, $\TemporalArchitecture$}\label{sub:longterm_reasoning}

To gather long-term temporal information, we use a 1-layer bidirectional Gated
Recurrent Unit (GRU~\cite{gatedrnn}) network $\TemporalArchitecture$, which
processes the dense per-frame features produced by $\FeatureExtractor$.
We set the hidden dimension of $\TemporalArchitecture$ to match that of $\FeatureExtractor$.
Finally, we apply a fully connected layer and softmax on the GRU outputs to make a per-frame $\NumClasses + 1$ way prediction (including 1 `no-event' background class).

We found that a single-layer GRU suffices and that more complex sequence models
such as MS-TCN~\cite{mstcn} or a deeper GRU do not necessarily improve accuracy
(see~\autoref{sub:ablation}).
We hypothesize that as a result of end-to-end training,
the features produced by $\FeatureExtractor$ capture subtle temporal cues
that are specific to a given activity's and task's requirements.
This shifts the burden of representations to \FeatureExtractor so
that \TemporalArchitecture only needs to propagate the temporal context.

\subsection{Per-frame Cross-Entropy Loss}
\label{sub:losses}

For a \NumFrames-frame clip, we output a sequence of \NumFrames class scores ---
i.e. a $(\NumClasses + 1)$-dimensional vector \Prediction{\Time} for each frame \Time, accounting for the background class:
\begin{equation}
    (\AllPredictions) = \TemporalArchitecture \circ \FeatureExtractor( \AllFrames ).
\end{equation}
Each frame has a ground-truth label $\Label{t} \in \{\AllClasses\} \cup \{ \Class{background} \}$ encoded as a one-hot vector.
We optimize per-frame classification with cross-entropy loss:
\begin{equation}
    l(\AllFrames) = \sum_{t=1}^{\NumFrames} \CE(\Prediction{t}, \Label{t})
    \label{eq:loss}
\end{equation}

\input{src/table/model_detail.tex}

\subsection{Implementation Details}
\label{sub:implementation}

We conduct experiments with two versions of \FeatureExtractor utilizing RegNet-Y 200MF and 800MF (MF refers to MFLOPs~\cite{regnet}).
These CNN backbones are initialized with pre-trained weights from ImageNet-1K~\cite{imagenet}.
Details of the complexity and throughput of these models is given in~\autoref{tab:model_throughput}.

We train \OURMETHOD on 100-frame-long clips sampled randomly and use standard data-augmentations (e.g., crop, jitter, and mixup~\cite{mixup}).
Frames are resized to 224 pixels in height and cropped to $224\times224$ unless otherwise stated (see~\autoref{sec:supp_our_impl}).
We optimize using AdamW~\cite{adamw} and LR annealing~\cite{cosinelr}.
To mitigate imbalance arising from the rarity of precise events ($<3\%$ of frames), we boost the loss weight of the foreground classes (5$\times$) relative to the background.

At test time, we disable data-augmentation and overlap clips by 50\%, averaging
the per-frame predictions.
To convert per-frame class scores into a set of spotting
predictions, we rank all of the frames by their predicted score for each class.
We follow standard procedure from coarse spotting~\cite{soccernetv2} and other detection
tasks~\cite{rcnn} by reporting our results with non-maximum suppression
(NMS).
Empirically, we found NMS's efficacy to vary by model and dataset (see~\autoref{tab:frame_accurate_results}).
Refer to~\autoref{sec:supp_our_impl} for more implementation details.

%% file: src/figure/pipeline.tex
\begin{figure}[t]
	\centering
	\includegraphics[width=\columnwidth]{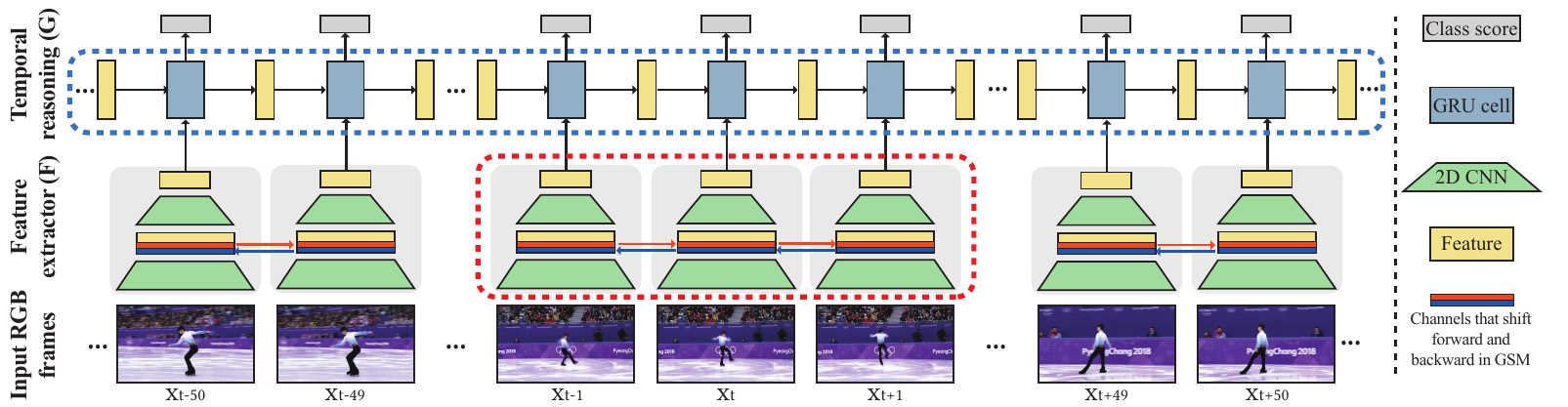}
	\caption{{\bf Overview of \OURMETHOD}.
		RGB video frames are first input to a local spatial-temporal feature extractor \FeatureExtractor (a RegNet-Y~\cite{regnet} with GSM~\cite{gsm}) to produce a feature for each frame that captures subtle differences and motion across neighboring frames (red dotted box).
		The feature sequence is then processed by a sequence model \TemporalArchitecture, which builds a long-scale temporal context (blue dotted box; one direction drawn) and outputs a class prediction for every frame.
		}
	\label{fig:pipeline}
\end{figure}

%% file: src/table/model_detail.tex
\renewcommand{\tabcolsep}{0.09cm}
\begin{table*}[t]
    \caption{{\bf \OURMETHOD efficiency and throughput.} We compare the model complexity, the maximum batch size for {\em end-to-end training on 100 frame clips} (at $224\times224$), and per-frame inference time on a Nvidia A5000 GPU with 24GB of VRAM~\cite{rtxa5000}.
    \OURMETHOD is significantly faster at inferring features than striding a video classification model and allows for practical end-to-end trained spotting.
    }
    \label{tab:model_throughput}
    {\scriptsize
    \centering
    \begin{tabularx}{\columnwidth}{lrrrr}
        \toprule
        \multicolumn{1}{c}{Architecture} & Params (M) & Max batch size & Inference time (ms) \\
        \midrule
        \OURMETHOD: RegNet-Y 200MF w/ GSM + GRU & (2.8 + 1.7) & 18 & 0.3 \\
        \OURMETHOD: RegNet-Y 800MF w/ GSM + GRU & (5.5 + 7.1) & 8 & 0.6 \\
        \midrule
        \midrule
        \multicolumn{4}{l}{\emph{Comparison to other feature extractors: (* $:=$ exceeds GPU memory)}} \\
        \midrule
        \multicolumn{2}{l}{RegNet-Y 200MF w/ GSM (7 frames per window)} 2.8 & 2 & 1.6 \\
        \multicolumn{2}{l}{RegNet-Y 200MF w/ GSM (15 frames per window)} 2.8 & 1 & 3.2 \\
        I3D (21 frames; used by~\cite{mstcn}) & 12.3 & * & 8.5 \\
        \multicolumn{2}{l}{R(2+1)D-34~\cite{r21d} (12 frames, $128\times128$; used by~\cite{tsp})} 63.7 & * & 11.0 \\
        ResNet-152 (1 frame only; used by~\cite{calf,netvladpp,rmsnet}) & 60.2 & 2 & 1.8 \\
        \multicolumn{2}{l}{Feature combination (for \soccernet)~\cite{featurecombattention}} $>$200 & - & - \\
        \bottomrule
    \end{tabularx}
    }
\end{table*}


%% file: src/dataset.tex
\section{Datasets}
\label{sec:dataset}

We evaluate precise spotting on four fine-grained sports video datasets with
frame-level labels: Tennis~\cite{vid2player}, Figure
Skating~\cite{vpd},~\finediving~\cite{finediving}, and~\finegym~\cite{finegym}.
For full details about these datasets, please refer to~\autoref{sec:supp_dataset}.

\textit{Tennis} is an extension of the dataset from Vid2Player~\cite{vid2player}.
It consists of 3,345 video clips from 28 tennis matches (each clip is a `point'), with video frame rates of either 25 or 30 FPS.
The dataset has 33,791 frame-accurate events divided into six classes:
``player serve ball contact,'' ``regular swing ball contact,'' and ``ball bounce'' (each divided by near- and far-court).
Video from 19 matches are used for training and validation, while 9 matches are held out for testing.

\textit{Figure Skating}~\cite{vpd} consists of 11 videos (all 25 FPS) containing 371 short program
performances from the Winter Olympics (2010--2018) and World Championships (2017--2019).
We refine the original labels by manually (re-)annotating the take-off and landing frames of jumps and flying spins, resulting in 3,674 event annotations across four classes.
We consider two splits for evaluation:
\begin{itemize}
    \item {\it Competition split} (\fscomp): holds out all videos from the 2018 season for testing~\cite{vpd}.
    This split tests generalization to new videos (e.g., the next Olympics), despite domain-shift such as a new background in a new venue.
    \item {\it Performance split} (\fsperf): stratifies each competition across train / val / test.
    This split tests a model's ability to learn precise temporal events (by different skaters) without the background bias of the previous split.
\end{itemize}

\textit{FineDiving}~\cite{finediving} contains 3,000 diving clips with temporal segment annotations.
We spot the step transition frames for four classes, which include transitions into somersaults (pike and tuck), twists, and entry.

\textit{FineGym}~\cite{finegym} contains 5,374 gymnastics performances, each treated as an untrimmed video.
It has 32 spotting classes, derived from a hierarchy of action categories
(e.g., balance beam dismounts; floor exercise turns).
The original annotations denote the start and end of actions;
we treat these boundaries as events --- for instance, ``balance beam dismount start'' and ``balance beam dismount end''.
We ignore the original splits, which are designed for action recognition and have overlap in videos, and we propose a 3:1:1 split between train / val / test.
To reduce the variation in the frame rates of the source videos (which are 25--60 FPS), we resample all 50 and 60~FPS videos to 25 and 30~FPS, respectively.

Upon inspecting the \finegym labels for frame accuracy, we found the annotations for action start frames to be more visually consistent than those for end frames.
For example, unlike in the Figure Skating dataset, the end frame is often several frames after the frame of landing for a jump.
Thus, we also report results for a subset, \finegym-Start, which contains only start-of-action events.

%% file: src/result.tex
\section{Evaluation}
\label{sec:result}

In~\autoref{sec:performance}, we demonstrate that the quality of per-frame feature representations extracted from the video has the
greatest impact on results, rather than the choice of head architecture, and that end-to-end learning with \OURMETHOD outperforms methods using pre-trained or indirectly fine-tuned features.
In~\autoref{sub:ablation} and~\autoref{sub:variations} we analyze the effect of temporal context, the importance of temporal modeling, and additional variations of \OURMETHOD.
In ~\autoref{sub:coarse_spotting} we report results on \soccernet, a temporally coarser spotting task.

\subsubsection*{Evaluation Metric.}
We measure Average Precision within a tolerance of \Tolerance frames (AP @ \Tolerance).
AP is computed for each event class, and mAP is the mean across classes.
We focus on tight tolerances such as $\Tolerance=1$ and $\Tolerance=2$.
Precise temporal events are rare as a percentage of frames (0.2--2.9\%), so
metrics such as frame-level accuracy are not meaningful for precise spotting.

\subsubsection*{Baselines.}
We evaluate \OURMETHOD against recent baselines from TAS, TAD, and coarse spotting that we adapted to the precise spotting task.
These methods are not trained end-to-end;
they adopt a two-phase separation between feature extraction and head training (i.e., downstream model) for the end-task.
We form our baselines by pairing a \emph{feature extraction strategy} with a \emph{spotting head}. The latter is trained on extracted features
to perform precise spotting, using the per-frame loss from~\autoref{eq:loss}.
See~\autoref{sec:supp_baseline_impl} for implementation details.

The baselines use the following head architectures: MS-TCN~\cite{mstcn}, GRU~\cite{gatedrnn}, ASFormer~\cite{asformer} from TAS; GCN~\cite{gtad} from TAD; and NetVLAD++~\cite{netvladpp} and transformer~\cite{featurecombattention} from action spotting.
MS-TCN, GRU, and ASFormer performed best in our experiments, so we relegate results from the remaining architectures to~\autoref{sub:supp_result_table}.
We further attempt to boost the performance of these baselines using additional losses from the spotting literature, such as CALF~\cite{calf} and label dilation\footnote{Label dilation is defined as naive propagation to $[-1,+1]$ frames to mitigate sparsity.}, and by post-processing using non-maximum suppression (within $\pm1$ frames).
We report results from the best configuration of each baseline.

We pair each head architecture with pre-extracted input features, grouped into three broad categories:
\begin{enumerate}
	\item {\em Pre-trained features} from video classification on Kinetics-400~\cite{kinetics}, which are often used without any fine-tuning for TAD and TAS.
	Like Farha et al.~\cite{mstcn}, we extract per-frame I3D features by densely striding a 21-frame window around each frame.
	To test the impact of better pre-trained models, we also extract features with MViT-B~\cite{mvit}, a state-of-the-art model from 2021.
	\item {\em Fine-tuned features} using TSP~\cite{tsp} and (\NumClasses + 1)-way clip classification\footnote{For direct comparison, (\NumClasses + 1)-VC uses the same RegNet-Y 200MF w/ GSM CNN backbone as \OURMETHOD. See~\autoref{sec:supp_baseline_impl} for details.}.
	These features come from a classifier trained to predict whether a small window (e.g., 12 frames) contains an event, and they have the benefit of being adapted to the target video domain (e.g., tennis, skating, gymnastics).
	\item {\em Pose features} (VPD) for the Figure Skating dataset only, which utilize a hand-engineered pipeline for subject tracking and fine-tuning~\cite{vpd}.
	\emph{These features utilize domain-knowledge and are costly to develop for new datasets}, which may include phenomena not captured by pose (e.g., ball bounce in tennis).
	In activities such as figure skating, defined heavily by human motion, VPD features serve as a ceiling for \OURMETHOD, which is domain agnostic.
\end{enumerate}

Finally, we add a naive, end-to-end learned baseline that adapts video classification directly to the spotting task (VC-Spot).
VC-Spot is given a 15-frame clip and tasked to predict whether the middle frame is a precise event.
This baseline is to show that precise spotting is a distinct task from video classification.

\input{src/table/frame_accurate.tex}
\input{src/table/ablation.tex}

\subsection{Spotting Performance}
\label{sec:performance}

We present two variations of \OURMETHOD in the main results: (1) a default configuration with a RegNet-Y~\cite{regnet} 200MF CNN backbone and RGB input only, and (2) a configuration using RegNet-Y 800MF with RGB and flow input.

\OURMETHOD with a 200MF CNN and RGB inputs consistently outperforms all non-pose baselines, while being comparable to the pose ones.
The benefits of \OURMETHOD are most striking at the most stringent
tolerance, $\Tolerance=1$ frame (\autoref{tab:frame_accurate_results}e).
We summarize the key takeaways of our evaluation below.

\textit{Pre-trained features generalize poorly} when no
fine-tuning is used, regardless of the head architecture: between 9.1--29.1 worse than \OURMETHOD in mAP at $\Tolerance=1$ (\autoref{tab:frame_accurate_results}a).
\textit{Fine-tuning yields a significant improvement} over pre-trained features: between 3.9--25.1 mAP at $\Tolerance=1$ (\autoref{tab:frame_accurate_results}b), indicating a large domain gap between Kinetics and the fine-grained spotting datasets.
However, \textit{\OURMETHOD further outperforms the two-phase approaches with fine-tuned features} by 3.3--6.8 mAP, showing that indirect fine-tuning strategies for temporal localization tasks should be compared against directly supervised, end-to-end learned baselines.
Finally, the wide variation in baseline performance (by sport) highlights the importance of evaluating new tasks, such as precise spotting, and their methods on a visually and semantically diverse set of activities and datasets.

\textit{VC-Spot performs poorly} compared to \OURMETHOD (\autoref{tab:frame_accurate_results}d), especially on Figure Skating and \finegym, which require temporal understanding at longer timescales (e.g., several seconds) compared to Tennis and \finediving.

\textit{\OURMETHOD achieves similar results to pose features} (2D-VPD~\cite{vpd}) on Figure Skating, within 0.1--2.5 mAP at $\Tolerance=1$.
This is encouraging because \OURMETHOD assumes no domain knowledge and is a more generally applicable approach.

\autoref{tab:frame_accurate_results}e also shows \OURMETHOD's \textit{best configuration}, using the larger 800MF CNN and both RGB and flow~\cite{raft}.
Neither of these enhancements (e.g., a larger CNN or flow) require domain knowledge, but can provide a small boost to the final performance over our 200MF defaults (0.8 mAP on Tennis and 3.9--4.3 mAP on \finegym).
Details for other \OURMETHOD configurations are presented in~\autoref{sub:variations}.

\subsection{Ablations of \OURMETHOD}
\label{sub:ablation}

We analyze the requirements of precise spotting with respect to temporal context and network architecture. Refer to~\autoref{sec:supp_result} for additional ablations.

\subsubsection*{Sensitivity to Clip Length.}
As a sequence model, \OURMETHOD can benefit from and make stateful predictions over a long temporal context (e.g., 100s of frames).
A long clip length allows for greater temporal context
for each prediction, but linearly increases memory utilization per batch.
We consider the number of frames needed for peak accuracy and train~\OURMETHOD with different clip lengths.
\autoref{tab:ablation_ourmethod}a shows that different activities require
different amounts of temporal context; the fast-paced events in Tennis can be
successfully detected even when context is only 8--16 frames.
In contrast, Figure Skating and \finegym show a clear drop in performance when clip length is reduced from 100 frames.
Even longer clip lengths may be desirable (e.g., 250 frames), though with diminishing returns.

\subsubsection*{Value of Temporal Information in the Per-frame Features.}
\OURMETHOD incorporates temporal information both in the 2D CNN backbone
\FeatureExtractor (with GSM) and after global spatial-pooling in
\TemporalArchitecture (with GRU).
We show the criticality of both of these components in
\autoref{tab:ablation_ourmethod}b at $\Tolerance=1$.
With neither GSM nor the GRU, the spotting task becomes a single-image
classification problem; as expected, the results are poor (at least $-21$ mAP).
The best results are achieved with both GSM and the GRU, except on \fsperf and \finediving, where results with and without GSM are similar.
Replacing GSM with TSM~\cite{tsm} (fixed shift) ranges from comparable to worse, showing GSM to be a reasonable starting default.

\subsubsection*{Spatial Resolution.}
Lowering spatial resolution~\cite{afsd,e2etad} can speed up end-to-end learning and inference but degrades mAP on precise spotting (\autoref{tab:ablation_ourmethod}c), where the subjects may, at times, occupy only a small portion of the frame.

\subsection{Additional Variations of \OURMETHOD}
\label{sub:variations}

\subsubsection*{More Complex Architectures, $\TemporalArchitecture$.}
Prior TAD and TAS works catalog a rich history of head architectures (see related;~\autoref{sec:related}) operating on pre-extracted features.
We examine whether these architectures can directly benefit from end-to-end learning with~\OURMETHOD by replacing the 1-layer GRU.
\autoref{tab:ablation_ourmethod}d shows that improvement is not guaranteed; MS-TCN, ASFormer, and deeper GRUs neither consistently nor significantly outperform a single layer GRU.
This suggests that \emph{end-to-end learned spatial-temporal features can already capture much of the logic previously handled by the downstream architecture.}

\subsubsection*{Enhancements to Feature Extractor, $\FeatureExtractor$.}
We explore two basic enhancements to $\FeatureExtractor$ that do not require new assumptions or domain knowledge: a larger CNN backbone (such as RegNet-Y 800MF) and optical flow~\cite{raft} input.
\autoref{tab:ablation_ourmethod}e shows that these enhancements can yield modest improvements (up to 4.4 mAP on \finegym).
Flow, by itself, is worse than RGB but can improve results when ensembled with RGB.
Larger models show promise on some datasets, but
the improvements are not as significant as the lift from end-to-end learning.

\subsection{Results on the SoccerNet Action Spotting Challenge}
\label{sub:coarse_spotting}

\input{src/table/soccernet.tex}

\OURMETHOD also generalizes to temporally coarse spotting tasks, such as \soccernet~\cite{soccernetv2}, which studies 17 action classes in 550 matches --- split across train / val / test / challenge sets.
As in prior work~\cite{calf,netvladpp,rmsnet}, we extract frames at 2 FPS and evaluate using average-mAP across tolerances, defined as $\pm \Tolerance / 2$ second ranges around events.
In~\autoref{tab:soccernet_results}, we compare \OURMETHOD to the best results from the CVPR 2021 (lenient tolerances of 5--60 sec) and CVPR 2022 (less coarse, 1--5 sec tolerances) SoccerNet Action Spotting challenges~\cite{snspotting}.

\OURMETHOD, with the 200MF CNN, matches the
top prior method from the 2021 competition~\cite{featurecombattention} in the 5--60 sec setting while outperforming it by 13.7--14.1 avg-mAP points in the less coarse, 1--5 sec setting.
Increasing the CNN to 800MF improves avg-mAP slightly (by 0.4--2.7 avg-mAP).
\OURMETHOD places second in the (concurrent) 2022 competition (within 1.1 avg-mAP), after Soares et al.~\cite{densedetectionanchorsrevisited}, due to the latter's strong performance on unshown actions (not visible in the frame).
Soares et al.~\cite{densedetectionanchors,densedetectionanchorsrevisited} and Zhou et al.~\cite{featurecombattention} are two-phase approaches, combining pre-extracted features from multiple (5 to 6) heterogeneous, fine-tuned feature extractors and proposing downstream architectures and losses on those features.
In contrast, \OURMETHOD shows that direct, end-to-end training of a simple and compact model can be a surprisingly strong baseline.

%% file: src/table/frame_accurate.tex
\renewcommand{\tabcolsep}{0.05cm}
\begin{table*}[t]
    \caption{{\bf Spotting performance (mAP @ $\delta$ frames).}
    The top results in each category and each column are \best{underlined}.
    SOTA is \sota{bold}.
    We report best results under the following: \textdagger~indicates NMS; * indicates CALF~\cite{calf} or dilation.
    (e) \OURMETHOD, trained with RGB only, generally outperforms the non-pose baselines and is competitive with the pose baselines on Figure Skating.
    \OURMETHOD can further be improved with a larger 800MF CNN and a 2-stream ensemble with flow.
    }
    \label{tab:frame_accurate_results}
    {
    \tiny
    \centering
    \begin{tabularx}{\textwidth}{ll
        rr
        rr
        rr
        rr
        rr
        rr
    }
        \toprule
        &
            & \multicolumn{2}{c}{\tennis}
            & \multicolumn{2}{c}{\fscomp}
            & \multicolumn{2}{c}{\fsperf}
            & \multicolumn{2}{c}{\finediving}
            & \multicolumn{2}{c}{FG-Full}
            & \multicolumn{2}{c}{FG-Start}
            \\
        \multicolumn{1}{c}{Feature} & \multicolumn{1}{c}{Model}
            & \multicolumn{1}{c}{$\delta$=1}
            & \multicolumn{1}{c}{2}
            & \multicolumn{1}{c}{1}
            & \multicolumn{1}{c}{2}
            & \multicolumn{1}{c}{1}
            & \multicolumn{1}{c}{2}
            & \multicolumn{1}{c}{1}
            & \multicolumn{1}{c}{2}
            & \multicolumn{1}{c}{1}
            & \multicolumn{1}{c}{2}
            & \multicolumn{1}{c}{1}
            & \multicolumn{1}{c}{2}
            \\
        \midrule
        \multicolumn{12}{l}{\em(a) Pre-trained features (from Kinetics-400)} \\
        I3D~\cite{i3d}
        & MS-TCN
            & 62.7 & \nms \loss75.4
            & 60.8 & \nms \loss79.5
            & \loss \best{69.0} & \nms\loss 89.3
            & - & -
            & - & -
            & - & -
            \\

        (RGB \& flow)
        & GRU
            & \nms \loss 45.7 & \nms \loss 70.5
            & \loss 41.8 & \nms \loss69.8
            & \loss 52.5 & \nms \loss77.5
            & - & -
            & - & -
            & - & -
            \\

        & ASFormer
            & \loss 58.1 & \nms\loss 76.5
            & \loss \best{61.2} & \nms\loss \best{82.4}
            & \best{69.0} & \nms\loss \best{89.7}
            & - & -
            & - & -
            & - & -
            \\

        MViT-B~\cite{mvit}
        & MS-TCN
            & \best{67.0} & \nms \loss 80.1
            & \loss 57.4 & \nms \loss 79.9
            & \loss 64.8 & \nms \loss 84.3
            & \loss \best{59.3} & \nms \loss \best{78.3}
            & \nms \best{31.0} & \nms \loss \best{48.6}
            & \nms \best{41.7} & \nms \loss \best{64.8}
            \\

        (RGB)
        & GRU
            & 64.8 & \nms \loss \best{80.8}
            & 45.6 & \nms \loss 73.1
            & 56.8 & \nms \loss 79.1
            & 57.3 & 76.7
            & \nms \loss 28.5 & \nms \loss \best{48.6}
            & \nms \loss 39.1 & \nms \loss 62.2
            \\

        & ASFormer
            & \loss 63.9 & \nms 79.9
            & 55.8 & \nms \loss  81.8
            & \loss 56.5 & \nms \loss 81.7
            & \loss 38.5 & \nms \loss 67.4
            & \nms \loss 25.3 & \nms \loss 42.9
            & \nms \loss 32.5 & \nms \loss 55.3 \\

        \midrule
        \multicolumn{12}{l}{\em (b) Fine-tuned features} \\
        TSP~\cite{tsp}
        & MS-TCN
            & \loss 90.9 & \nms \loss95.1
            & 72.4 & \nms \loss 87.8
            & \loss76.8 & \loss 89.9
            & \loss 57.7 & \nms 76.0
            & \nms 40.5 & \nms 58.5
            & \nms 53.9 & \nms \loss 73.5
            \\

        (RGB)
        & GRU
            & 89.5 & \nms \loss 96.0
            & \loss 68.4 & \nms \loss 88.3
            & 75.5 & \nms \loss 90.6
            & \loss 57.0 & \loss 78.2
            & \nms \loss 38.7 & \nms \loss \best{58.8}
            & \nms \loss 53.2 & \nms \loss \best{74.2}
            \\

        & ASFormer
            & 89.8 & \nms\loss 95.5
            & \best{77.7} & \nms \best{94.1}
            & \best{80.2} & \nms \best{94.5}
            & \loss 51.3 & \nms \loss 77.4
            & \nms 38.8 & \nms 57.6
            & \nms 51.1 & \nms\loss 72.9
            \\

        (\NumClasses + 1)-VC
        & MS-TCN
            & 91.1 & \nms \loss 95.1
            & 66.5 & \nms 77.2
            & \loss 77.2 & \nms \loss 89.9
            & \best{63.2} & \nms \loss \best{83.5}
            & \nms 40.9 & \nms \loss 58.2
            & \nms 53.2 & \nms \loss 73.8
            \\

        (RGB)
        & GRU
            & \nms \loss 91.5 & \nms \loss \best{96.2}
            & \nms \loss 61.7 & \nms \loss 78.9
            & \nms \loss 76.8 & \nms \loss 89.4
            & \loss 61.8 & \nms \loss 82.6
            & \nms \best{41.1} & \nms 57.9
            & \nms \best{54.3} & \nms \loss 73.6
            \\

        & ASFormer
            & \best{92.1} & \nms\loss \best{96.2}
            & \loss 67.6 & \nms\loss 79.8
            & 77.1 & \nms\loss 89.8
            & \loss 58.9 & \nms\loss \best{83.5}
            & \nms 40.0 & \nms\loss 56.9
            & \nms\loss 53.6 & \nms\loss 72.9
            \\
        \midrule
        \multicolumn{12}{l}{\em (c) Hand-engineered tracking \& pose features (top scores shown; see~\autoref{sub:supp_result_table} for GRU and ASFormer)} \\
        2D-VPD~\cite{vpd}
        & MS-TCN
            & - & -
            & \loss \sota{83.5} & \nms \loss \sota{96.2}
            & \loss \sota{85.2} & \nms \loss \sota{96.4}
            & - & -
            & - & -
            & - & -
            \\

        \midrule
        \multicolumn{12}{l}{\em (d) VC-Spot: video classification baseline using RGB} \\
        \multicolumn{2}{l}{RegNet-Y 200MF w/ GSM}
            & \nms \best{92.4} & \nms 96.0
            & \nms 61.8 & \nms 75.5
            & \nms 56.2 & \nms 75.3
            & \nms 62.4 & \nms \sota{85.6}
            & \nms 18.7 & \nms 28.6
            & \nms 25.9 & \nms 38.3
            \\

        \midrule
        \multicolumn{12}{l}{\em (e) \OURMETHOD}\\
        \multicolumn{2}{l}{{\bf Default:} 200MF (RGB)}
            & 96.1 & \nms 97.7
            & \nms \loss 81.0 & \nms \loss 93.5 
            & \nms \loss\best{85.1} & \nms \loss 95.7 
            & \sota{68.4} & \nms \best{85.3}
            & \nms 47.9 & \nms 65.2
            & \nms 61.0 & \nms 78.4
            \\
        \multicolumn{2}{l}{{\bf Best}: 800MF (2-stream)}
            & \nms \sota{96.9} & \nms \sota{98.1}
            & \nms \loss \best{83.4} & \nms \loss \best{94.9} 
            & \nms \loss 83.3 & \nms \loss \best{96.0} 
            & \nms 66.4 & \nms 84.8
            & \nms \sota{51.8} & \nms \sota{68.5}
            & \nms \sota{65.3} & \nms \sota{81.6}
            \\

        \bottomrule
    \end{tabularx}
    }
\end{table*}

%% file: src/table/ablation.tex
\newcommand{\tabindent}[0]{\,\,\,\,}
\renewcommand{\tabcolsep}{0.105cm}
\begin{table*}[t]
    \caption{{\bf Ablation and analysis experiments (mAP @ $\delta=1$).}
    We compare to \OURMETHOD defaults in the top row (RegNet-Y 200MF w/ GSM and GRU).
    (a) Varying clip lengths show that temporal context from longer clips is generally helpful.
    (b) Removing temporal information in the feature extractor $\FeatureExtractor$ (GSM) and in the stateful predictions $\TemporalArchitecture$ (GRU) generally reduces mAP.
    (c) Reducing input resolution from 224 to 112 pixels reduces mAP.
    (d) More complex models for $\TemporalArchitecture$ than the 1-layer GRU do not significantly improve mAP.
    (e) Enlarging $\FeatureExtractor$ to 800MF and/or adding flow can improve mAP slightly on some datasets.}
    \label{tab:ablation_ourmethod}
    {
    \tiny
    \centering
    \begin{tabularx}{\textwidth}{ll rr rr rr rr rr}
        \toprule
        & & \multicolumn{2}{c}{\tennis}
            & \multicolumn{2}{c}{\fscomp}
            & \multicolumn{2}{c}{\fsperf}
            & \multicolumn{2}{c}{\finediving}
            & \multicolumn{2}{c}{\finegym-Full} \\
        \multicolumn{2}{c}{Experiment}
            & \multicolumn{1}{c}{mAP} & \multicolumn{1}{c}{$\Delta$}
            & \multicolumn{1}{c}{mAP} & \multicolumn{1}{c}{$\Delta$}
            & \multicolumn{1}{c}{mAP} & \multicolumn{1}{c}{$\Delta$}
            & \multicolumn{1}{c}{mAP} & \multicolumn{1}{c}{$\Delta$}
            & \multicolumn{1}{c}{mAP} & \multicolumn{1}{c}{$\Delta$} \\
        \midrule
        \multicolumn{2}{l}{\OURMETHOD default: clip length = 100}
            & 96.1 &
            & \nms 81.0 &
            & \nms 85.1 &
            & 68.4 &
            & \nms 47.4 & \\
        \midrule
        (a)
        & \tabindent clip length = 8
            & \nms 95.8 & $-0.3$
            & \nms 73.7 & $-7.3$
            & \nms 74.7 & $-10.4$
            & \nms 67.3 & $-1.1$
            & \nms 32.3 & $-15.1$ \\
        & \tabindent clip length = 16
            & \nms 96.2 & $+0.1$
            & \nms 74.4 & $-6.6$
            & \nms 80.1 & $-5.0$
            & \nms 64.8 & $-3.6$
            & \nms 40.8 & $-6.6$ \\
        & \tabindent clip length = 25
            & \nms 96.2 & $+0.1$
            & \nms 74.5 & $-6.5$
            & \nms 80.6 & $-4.5$
            & \nms 67.2 & $-1.2$
            & \nms 43.9 & $-3.5$ \\
        & \tabindent clip length = 50
            & \nms 96.4 & $+0.3$
            & \nms 76.9 & $-4.1$
            & \nms 82.3 & $-2.8$
            & 65.0 & $-3.4$
            & \nms 46.6 & $-0.8$ \\
        & \tabindent clip length = 250
            & 96.4 & $+0.3$
            & \nms 81.3 & $+0.3$
            & \nms 85.6 & $+0.5$
            & 68.9 & $+0.5$
            & \nms 48.5 & $+1.1$ \\
        & \tabindent clip length = 500
            & 95.9 & $-0.2$
            & \nms 78.9 & $-2.1$
            & \nms 87.5 & $+2.4$
            & - & -
            & \nms 48.1 & $+0.7$\\
        \midrule
        (b)
        & \tabindent w/o GRU
            & \nms 95.7 &  $-0.4$
            & \nms 74.3 & $-6.7$
            & \nms 77.9 & $-7.2$
            & 64.1 & $-4.3$
            & \nms 32.9 & $-14.5$ \\
        & \tabindent w/ TSM~\cite{tsm} instead of GSM
            & 96.1 & $+0.0$
            & \nms 78.6 & $-2.4$
            & \nms 83.3 & $-1.8$
            & \nms 65.3 & $-3.1$
            & \nms 48.1 & $+0.7$\\
        & \tabindent w/o GSM
            & \nms 94.1 & $-2.0$
            & \nms 75.5 & $-5.5$
            & \nms 85.6 & $+0.4$
            & 68.9 & $+0.5$
            & \nms 44.2 & $-3.2$ \\
        & \tabindent w/o GSM \& GRU
            & \nms 60.1 & $-36.0$
            & \nms 26.9 & $-54.1$
            & \nms 41.1 & $-44.0$
            & \nms 47.0 & $-21.4$
            & \nms 22.1 & $-25.3$ \\
        \midrule
        (c)
        & \tabindent w/ 112 px resolution (height)
            & \nms 88.5 & $-7.6$
            & \nms 75.4 & $-5.6$
            & \nms 80.9 & $-4.2$
            & \nms 64.9 & $-3.5$
            & \nms 45.3 & $-2.6$ \\
        \midrule
        (d)
        & \tabindent w/ MS-TCN
            & 95.7 &  $-0.4$
            & \nms 77.6 & $-3.4$
            & \nms 84.7 & $-0.4$
            & 67.0 & $-1.4$
            & \nms 44.1 & $-3.3$ \\
        & \tabindent w/ ASFormer
            & 95.7 &  $-0.4$
            & \nms 68.4 & $-12.6$
            & \nms 75.4 & $-9.7$
            & 70.4 & $+2.0$
            & \nms 36.8 & $-10.6$\\
        & \tabindent w/ Deeper GRU
            & 96.5 &  $+0.4$
            & \nms 80.2 & $-0.8$
            & \nms 83.5 & $-1.6$
            & 67.2 & $-1.2$
            & \nms 46.4 & $-1.0$ \\
        & \tabindent w/ GRU* (see supplement)
            & 96.2 &  $+0.1$
            & \nms 78.1 & $-2.9$
            & \nms 86.0 & $+0.9$
            & 67.4 & $-1.0$
            & \nms 47.9 & $+0.5$ \\
        \midrule
        (e)
        & 200MF (Flow)
            & \nms 58.2 & $-37.9$
            & \nms 72.4 & $-8.6$
            & \nms 76.6 & $-8.5$
            & \nms 60.7 & $-7.7$
            & \nms 44.4 & $-3.0$ \\
        & 200MF (RGB + flow; 2-stream)
            & \nms 96.3 & $+0.2$
            & \nms 82.2 & $+1.2$
            & \nms 85.1 & $+0.0$
            & \nms 70.1 & $+1.7$
            & \nms 49.0 & $+1.6$ \\
        \cmidrule{2-12}
        & 800MF (RGB)
            & 96.8 &  $+0.7$
            & \nms 84.0 & $+3.0$
            & \nms 83.6 & $-1.5$
            & 64.6 & $-3.8$
            & \nms 50.1 & $+2.7$ \\
        & 800MF (Flow)
            & \nms 59.2 & $-36.9$
            & \nms 74.9 & $-6.1$
            & \nms 74.2 & $-10.9$
            & \nms 59.8 & $-8.6$
            & \nms 46.9 & $-0.5$ \\
        & 800MF (RGB + flow; 2-stream)
            & \nms 96.9 & $+0.8$
            & \nms 83.4 & $+2.4$
            & \nms 83.3 & $-1.8$
            & \nms 66.4 & $-2.0$
            & \nms 51.8 & $+4.4$ \\
        \bottomrule
    \end{tabularx}
    }
\end{table*}

%% file: src/table/soccernet.tex
\renewcommand{\tabcolsep}{0.055cm}
\begin{table*}[t]
    \caption{{\bf Average-mAP @ $t$ for tolerances in seconds.}
    SOTA in \sota{bold}.
    We show the top results from the CVPR 2021 and 2022 SoccerNet Action Spotting challenges.
    \textdaggerdbl~indicates challenge results --- trained on the train, validation, and test splits.
    Shown and unshown refer to whether actions are visible;~\OURMETHOD is better at detecting the former, but Soares et al.~\cite{densedetectionanchorsrevisited} is superior at the latter.
    }
    \label{tab:soccernet_results}
    {
    \scriptsize
    \centering
    \begin{tabularx}{\textwidth}{lrrr|rr}
        \toprule
        & \multicolumn{2}{c}{Test split}
        & \multicolumn{3}{c}{Challenge split} \\
        \multicolumn{1}{c}{Average-mAP @ tolerances}
            & Tight (1--5 s)
            & Loose (5--60 s)
            & Tight (1--5 s)
            & Shown
            & Unshown \\
        \midrule
        RMS-Net~\cite{rmsnet}
            & 28.83
            & 63.49
            & 27.69
            & -
            & - \\
        \multicolumn{2}{l}{NetVLAD++~\cite{netvladpp}}
            - & - & 43.99 & - & - \\
        Zhou et al.~\cite{featurecombattention} (2021 challenge; 1st)
            & 47.05
            & 73.77
            & 49.56
            & 54.42
            & 45.42 \\
        \multicolumn{2}{l}{\textsuperddagger Soares et al.~\cite{densedetectionanchorsrevisited} (2022 challenge; 1st)}
            -
            & -
            & \textsuperddagger\sota{67.81}
            & \textsuperddagger72.84
            & \textsuperddagger\sota{60.17} \\
        \midrule
        \multicolumn{1}{l}{\OURMETHOD 200MF}
            & 61.19 & 73.25 & 63.28 & 70.41 & 45.98 \\
        \multicolumn{1}{l}{\OURMETHOD 800MF}
            & 61.82 & 74.05 & 66.01 & 72.76 & 51.65 \\
        \multicolumn{2}{l}{\textsuperddagger\OURMETHOD 800MF (2022 challenge; 2nd)}
            - & - & \textsuperddagger 66.73 & \textsuperddagger \sota{74.84} & \textsuperddagger 53.21 \\
        \bottomrule
    \end{tabularx}
    }
\end{table*}

%% file: src/discussion.tex
\section{Discussion and Future Work}
\label{sec:discussion}

In this paper, we have presented a from-the-ground-up study of end-to-end feature learning for spotting in the temporally stringent setting.

\OURMETHOD is a simple baseline that obtains competitive or state-of-the-art performance on temporally precise (and coarser) spotting tasks, outperforming conventional approaches derived from related work on TAD and TAS (\autoref{sec:related}).
The secondary benefits we obtain from end-to-end learning are a simplified analysis pipeline, trained in a single phase under direct supervision, and the ability to use smaller, simpler models, without sacrificing accuracy on the frame-accurate task.
Methodological enhancements such as improved architectures (e.g., based on ViT~\cite{vit}) for feature extraction, training methodologies, head architectures, and losses that benefit from end-to-end learning are interesting research directions.
We hope that \OURMETHOD serves as a principled baseline for this future work.

Video understanding encapsulates a broad body of tasks, of which spotting frame-accurate events is a single example.
We consider it future work to analyze other tasks and their datasets, and we  anticipate situations where end-to-end learning alone may be insufficient:
e.g., when reliable priors such as pose are readily available, or when training data is limited or exhibits domain-shift in the pixel domain.
Learning to spot accurately with few or weak labels will accelerate the curation new datasets for more advanced, downstream video analysis tasks.

%% file: src/conclusion.tex
\section{Conclusion}

We have introduced temporally precise spotting in video, supported by four fine-grained sports datasets.
Many recent advances in TAD, TAS, and spotting trend towards increasingly complex models and processing pipelines, which generalize poorly for this strict, but practical setting.
\OURMETHOD shows that a few key design principles --- task-specialized
spatial-temporal features, reasoning over sufficient temporal context, and efficient end-to-end learning --- can go a long way for improving accuracy and simplifying solutions.

%% file: src/ack.tex
\subsubsection*{Acknowledgements.}
This work is supported by the National Science Foundation (NSF) under III-1908727, Intel Corporation, and Adobe Research.
We also thank the anonymous reviewers for their comments and feedback.

%% file: src/supp_impl.tex
\section{Implementation Details for \OURMETHOD}
\label{sec:supp_our_impl}

\subsection{Spatial-Temporal Feature Extractor, \FeatureExtractor}

As described in~\autoref{sec:method}, our feature extractor is a standard RegNet-Y~\cite{regnet} with Gate Shift Modules~\cite{gsm} (GSM) inserted.
GSM is applied at each residual block, to $\frac{1}{4}$ of the channels, rounded up to the nearest multiple of 4.
RegNet-Y 200MF and 800MF produce spatially-pooled features of dimension 368 and 768 respectively.

We choose RegNet-Y~\cite{regnet} over the more commonly used ResNet~\cite{resnet} family of 2D CNNs because the former is more recent and compact (RegNet-Y 200MF has 3.2M parameters vs. 11.7M parameters for ResNet-18), while exhibiting generally better performance on image classification benchmarks~\cite{timm}.
\OURMETHOD, however, can be implemented with any 2D CNN architecture.

\subsection{Long-term Temporal Reasoning Module, \TemporalArchitecture}

\TemporalArchitecture provides temporal reasoning on dense feature vectors, following the spatial pooling layer of \FeatureExtractor.
The details of \TemporalArchitecture are given in the paper in~\autoref{sub:longterm_reasoning}.
Here, we provide details for the additional variations of \OURMETHOD used in~\autoref{sub:variations}.

\textit{Deeper GRU} increases the number of GRU layers to 3.
\textit{MS-TCN and ASFormer} are described in~\autoref{sub:supp_baseline_models}.

\textit{GRU*} takes multiple 1-layer GRUs at different temporal granularities, in addition to the 1-layer GRU, to more directly aggregate information across wider contexts.
We use two temporal scales, 4 and 16, requiring two additional GRUs.
Each scale defines a temporal down-sampling of the clip length by a factor of the scale, $S$.
For each scale, all output features are first fed to a fully connected layer and ReLU. Then, the sequence of length $\NumFrames$ is divided into $\lceil \frac{\NumFrames}{S} \rceil$ non-overlapping windows, and max-pooling is performed in each window.
The $\lceil \frac{\NumFrames}{S} \rceil$ sequence is processed by the scale-specific GRU.
Finally, the outputs of each GRU, at each time scale, are up-sampled by repetition back to the full clip length $\NumFrames$ and concatenated for each time step $\Time$.

While these experiments do not cover the full breadth of architectures and settings available, we note that we did not observe major performance gains over the 1-layer GRU in applying these alternatives alongside end-to-end learning.

\subsection{Training Configuration}

We train \OURMETHOD using 100 frame long clips by default and a batch size of 8 clips.
Batches are formed by randomly sampling clips from the training videos.
We group every 625 training steps into a training cycle (i.e., a pseudo-epoch of 500K frames).
A single cycle runs in approximately 8.5 and 14 minutes on a single A5000 GPU~\cite{rtxa5000} for the 200MF and 800MF variants, respectively.
All variations of \OURMETHOD are trained for 50 cycles on the Tennis, Figure Skating, and \finediving datasets.
We train the 200MF model for 100 cycles and the 800MF model for 150 cycles on \finegym and \soccernet, due to the larger dataset sizes (see~\autoref{sec:supp_dataset}).
Training is performed with AdamW~\cite{adamw}, setting a base learning rate of $10^{-3}$, with 3 linear warmup cycles followed by cosine decay~\cite{cosinelr}.

\subsubsection*{Data Augmentations.}
We randomly apply color jitter, Gaussian blur, and mixup~\cite{mixup} during training.
On Tennis, Figure Skating, and \finegym, we also randomly crop the $398\times224$ frames to $224\times224$ pixels.
This crop only affects the width dimension, as cropping the height dimension can lead to precise events falling outside the visible field (e.g., the tennis court and player span the vertical dimension).
For \finediving~\cite{finediving}, we use the frames extracted by the original authors (256 pixels in the vertical dimension) and random crops of $224\times224$ pixels.
Finally, for \soccernet, we do not use random cropping because context such as the goal or the field boundary are often at the periphery of the frame.

For Figure Skating only (\fsperf and \fscomp), we use label dilation of $\pm1$ frames due to the very large imbalance between events and background frames (see~\autoref{sub:supp_fs_dataset}).
Label dilation is beneficial on Figure Skating for both \OURMETHOD and the baselines (see~\autoref{sub:supp_calf_and_dilate}).
Note that label dilation is not used during testing.

\subsubsection*{Non-maximum Suppression.}
We evaluated the model predictions with and without non-maximum suppression (NMS).
For the temporally precise datasets, we used a window of $\pm1$ frames whereas we use $\pm2$ frames at 2 FPS for \soccernet.
The efficacy of NMS in the temporally precise setting depends on the frame level tolerance, dataset, and model (see experiments in~\autoref{sub:supp_ablate_nms}), so the decision to apply NMS in practice should be made with application and task requirements in mind.

\subsection{Optical Flow Extraction, for Additional Experiments}

We use optical flow extracted by RAFT~\cite{raft} for the additional 2-stream experiments that we described in~\autoref{sub:variations}.
During preprocessing, we subtract the median flow value for each frame and clamp to a range of $[-20, +20]$ pixels.

\section{Implementation Details for Baselines}
\label{sec:supp_baseline_impl}

We adapt a number of published architectures from the action segmentation (TAS), detection (TAD), and spotting literature as baselines for temporally precise spotting and provide their key implementation details here.

\subsection{Models}
\label{sub:supp_baseline_models}

\subsubsection*{TCN and MS-TCN.}
We adapt the code from Farha et al.~\cite{mstcn}, using dilated temporal convolution networks.
Multiple stages typically improves results over a single stage TCN.
We use 3 TCN stages for our MS-TCN baselines and a depth of 5 layers for each stage.
Each layer has dimension of 256. Per-frame predictions are made with a fully connected layer that maps from 256 to $\NumClasses + 1$.

\subsubsection*{GRU.}
We use a bidirectional GRU~\cite{gatedrnn} with 5 layers and a dimension $H$ of 128.
Per-frame predictions are made with a fully connected layer, from $2H$ to $\NumClasses + 1$.

\subsubsection*{ASFormer.}
We use code and settings from the implementation by Yi et al.~\cite{asformer}.

\subsubsection*{GCN.}
We use the GCNeXt block architecture proposed by Xu et al.~\cite{gtad}, which produces a 256 dimensional feature encoding for each frame.
Per-frame predictions are made with a fully connected layer mapping from 256 to $\NumClasses + 1$.

\subsubsection*{StridedTransformer.}
We implement a transformer~\cite{pytorchtransformer} that operates on a window of per-frame features~\cite{featurecombattention}.
The model takes a consecutive clip of 31 features and positional encodings, and it predicts whether the center frame is one of the $\NumClasses$ events or not.

\subsubsection*{NetVLAD++}~\cite{netvladpp} is used similarly to the transformer described above.
We observe on precise spotting tasks that NetVLAD++ often fails to overcome the class imbalance between foreground events and background frames.
Reducing window size from 31 to 7 frames improves performance slightly, but overall performance remains poor and the StridedTransformer described above performs significantly better (see~\autoref{sub:supp_result_table}).

\subsubsection*{VC-Spot} is a end-to-end learned video classification baseline, which, given a clip of 15 consecutive RGB frames, predicts whether the middle frame is an event. We use the same RegNet-Y 200MF (with GSM) CNN backbone as \OURMETHOD.
Training VC-Spot using batches containing randomly sampled clips fails to overcome the large foreground / background frame imbalance.
This is a challenging problem since a window that contains a temporally precise event as its middle frame differs from its neighbors by only one frame in time.
To ameliorate this, we form batches with densely overlapped clips (4 sequentially) in addition to the batch size of 8.

\subsection{Pre-trained Features}
\label{sub:supp_pretrained_features}

We test I3D~\cite{i3d} and MViT base (MViT-B)~\cite{mvit} features trained on Kinetics-400~\cite{kinetics}, without fine-tuning.
I3D features are extracted following the example of Farha et al.~\cite{mstcn}, with RGB and flow.
MViT-B features use the 16x4 model in PyTorchVideo~\cite{pytorchvideo}.
Performance with these features is poor --- far below fine-tuned features such as TSP~\cite{tsp} (see~\autoref{tab:supp_full_pretrained} and~\ref{tab:supp_full_fine_tuned}).
Due to the high cost of feature extraction on large datasets with I3D and the poor spotting performance of downstream models trained using I3D features, we only extract MViT-B~\cite{mvit} features for \finediving and \finegym.

\subsection{Fine-tuned Features}

We test two fine-tuning strategies that use video clip classification in the target domain (i.e., the precise spotting dataset) as a fine-tuning step for temporal localization tasks.

\subsubsection*{Temporally Sensitive Pretraining (TSP).}
We use code from Alwassel et al.~\cite{tsp}, which pre-trains a R(2+1)D-34~\cite{r21d} model to encode spatial-temporal features.
The model is first initialized with weights from a model trained on Kinetics-400~\cite{kinetics}.
During fine-tuning, we use a clip length of 12 frames.
For the pre-trained global video feature (GVF), we use pre-extracted MViT-B~\cite{mvit} features (from~\autoref{sub:supp_pretrained_features}) as these serve a similar function to the frozen GVF in the original implementation.
We optimize the model using TSP until its validation loss and accuracy converges.

\subsubsection*{(\NumClasses + 1)-VC} pre-trains a RegNet-Y 200MF with GSM on a standard video classification task.
It is included to demonstrate a simpler fine-tuning baseline than TSP, using a feature extractor of comparable complexity and architecture to the one that we selected for~\OURMETHOD.

We initialize the RegNet-Y backbone with pre-trained weights learned on ImageNet-1K~\cite{imagenet}.
For fine-tuning, we use a clip length of 7 frames.
A small clip length is selected because the goal is to learn a localized, per-frame feature; downstream models for spotting will receive a long sequence of these features.
Clips of the $\NumClasses$ foreground classes contain a foreground event within a half clip length window in the clip center while background class clips do not.
We sample background clips randomly with 20\% probability during training.
The model is trained with a batch size of 16 clips and for 18.8K steps.
The best epoch is selected using validation accuracy.

\subsubsection*{Video Pose Distillation} (VPD)~\cite{vpd} features are available for the Figure Skating dataset and serve as a strong baseline / performance target for \OURMETHOD.

The VPD features are learned in an unsupervised manner over the entire video dataset (including the test videos, without access to action or event labels).
They make use of hand-engineered subject tracking, RGB pixels, and optical flow as inputs.
We test both 2D-VPD and (view-invariant) VI-VPD features.
The differences are subtle when applied to precise spotting, with 2D-VPD being better a majority of the time (see~\autoref{sub:supp_result_table}).

\subsection{Training Configuration (for Spotting)}

With the exception of VC-Spot (an end-to-end learned baseline), all of the baseline architectures described in~\autoref{sub:supp_baseline_models} operate in two phases, learning a spotting head on densely pre-extracted features.

We train the TCN, MS-TCN, GRU, ASFormer, and GCN models on randomly sampled, 500 frame long clips --- with a batch size of 50, a train-val cycle of 40 steps (1M frames), and for 50 cycles.
Updates are performed using AdamW~\cite{adamw} with a base learning rate of $10^{-3}$, linear warmup (3 cycles), and cosine annealing~\cite{cosinelr}.
The StridedTransformer and NetVLAD++~\cite{netvladpp} baselines make singular predictions on a window of frames.
We train these with a batch size of 100 clips, train-val cycles of 1,000 steps, and for 50 cycles.
We use the same AdamW~\cite{adamw} optimizer and LR schedule as the other models.
Validation mAP, computed at the end of every training cycle, is used for model selection.

%% file: src/supp_result.tex
\section{Additional Experiments \& Ablations}
\label{sec:supp_result}

In \autoref{sub:supp_result_table} and~\autoref{sub:supp_calf_and_dilate}, we present additional baselines omitted from the main paper due to space constraints.
\autoref{sub:supp_ablate_nms} assesses the necessity of non-maximum suppression (NMS) for temporally precise spotting.
\autoref{sub:supp_result_exact} provides results when evaluating spotting performance at tolerance $\Tolerance=0$ frames (i.e., the exact frame of human annotation).
\autoref{sub:supp_breakdown_class} analyzes the variation in precise spotting performance among the event classes in each dataset.

\input{src/supp/tab_full_result.tex}

\input{src/supp/tab_nms_result.tex}

\input{src/supp/tab_frame_exact.tex}

\subsection{Full Baseline Result Tables}
\label{sub:supp_result_table}

We report the top baseline results in~\autoref{sec:performance}.
\autoref{tab:supp_full_pretrained},~\ref{tab:supp_full_fine_tuned}, and~\ref{tab:supp_full_result_pose} provide full results for all of the baselines and feature combinations.

For the best performing MS-TCN~\cite{mstcn}, GRU~\cite{gatedrnn}, and ASFormer~\cite{asformer} configurations, we further trained the model with and without CALF~\cite{calf} and label dilation (propagating labels to $\pm 1$ adjacent frames).
NetVLAD++~\cite{netvladpp} failed to overcome label sparsity in all tested datasets except for Tennis (with fine-tuned features).
The StridedTransformer~\cite{pytorchtransformer} performed better than NetVLAD++ and was tested with and without label dilation ($\pm 1$ frames), as it also suffers from sparsity in the foreground labels.

\subsection{Impact of Additional Losses on Baseline Performance}
\label{sub:supp_calf_and_dilate}

Losses such as CALF~\cite{calf} have been proposed in spotting literature as a way to address sparsity in temporal event labels.
In the interest of obtaining strong baselines for precise spotting, we attempt to boost the top performing model architecture and feature baselines in~\autoref{sub:supp_result_table}.

We add CALF as an additional loss, with parameters that smooth around a event within a 7 frame window.
Conceptually, because of the tight tolerances in temporally precise spotting, the small number of frames in an appropriately sized window prevents the loss from achieving as smooth as an effect as in coarse action spotting.
We also implemented a simpler label dilation baseline, which addresses the sparsity problem by propagating event labels to $\pm 1$ frame before and after each event at training time (denoted as ``dilate 1'').

\autoref{tab:supp_full_pretrained},~\ref{tab:supp_full_fine_tuned}, and \ref{tab:supp_full_result_pose} list results with CALF and label dilation for the MS-TCN~\cite{mstcn}, GRU~\cite{gatedrnn}, and ASFormer~\cite{asformer} architectures.
The results are generally mixed, with scores being similar with and without these loss modifications (e.g., within 1-2 mAP @ $\Tolerance=1$).
On \fscomp, the difference is more pronounced with 2D-VPD~\cite{vpd} features --- up to 6.3 mAP improvement.

\subsection{Sensitivity of Results to Non-Maximum Suppression}
\label{sub:supp_ablate_nms}

Non-maximum suppression (NMS) is a common post-processing technique in detection tasks~\cite{soccernetv2,rcnn}.
We find that, for precise spotting, NMS is typically beneficial at tolerances of $\Tolerance\geq2$ frames but may be harmful for $\Tolerance \leq 1$ frame (see~\autoref{tab:supp_nms_result}).
Tuning the NMS window threshold past 1 frame often has a minimal effect of less than 1 mAP point.

\subsection{Predicting the Exact Frame of Human Annotation}
\label{sub:supp_result_exact}

While our spotting datasets have annotations at the frame-level, the $\Tolerance=0$ frame-prediction task is especially challenging to scientifically evaluate.
In 25--30 FPS video, quick events such as a ``ball bounce'' can fall between two adjacent frames.
$\Tolerance=0$ is also unforgiving of any small inconsistencies in labeling.
Ignoring these limitations, \OURMETHOD outperforms the baseline approaches, and compares similarly to models using hand-engineered pose features, in agreement with human annotators (\autoref{tab:supp_result_exact}).
The practical meaning of mAP @ $\Tolerance=0$, however, is limited due to the aforementioned confounds.

\subsection{Visualizing the Spotting Performance of Different Classes}
\label{sub:supp_breakdown_class}

The difficulty of precisely spotting events can vary by event class.
In~\autoref{fig:supp_pr_curves}, we show interpolated precision-recall curves for the different classes in the Tennis, Figure Skating, \finediving, and \finegym datasets from our default \OURMETHOD 200MF model trained on RGB inputs.

While spotting performance is similar among the different classes that comprise Tennis, Figure Skating, and \finediving,
spotting on \finegym shows a large amount of variation; some classes such as
``balance beam dismounts start'' and ``floor exercise front\_salto start''
are spotted with high precision and recall at $\Tolerance=1$, while other classes such as ``vault (timestamp 0)'' and ``balance beam turns end'' exhibit much lower performance.
We noted in~\autoref{sec:dataset} that there is variation in the visual precision of different \finegym classes, where the annotated frames do not necessarily map to salient visual events.

\input{src/supp/fig_pr_curve.tex}

%% file: src/supp/tab_full_result.tex
\renewcommand{\tabcolsep}{0.05cm}
\begin{table*}[p]
    \centering
    \caption{{\bf Spotting performance (mAP @ $\delta$ frames)} using pre-trained features without fine-tuning.
    \textdagger~indicates NMS.
    The best baseline scores are \best{underlined}.
    Due to the low performance of I3D~\cite{i3d} features (compared to TSP~\cite{tsp}), we do not extract I3D features for \finediving and \finegym.
    }
    \label{tab:supp_full_pretrained}
    {
    \tiny
    \begin{tabularx}{\textwidth}{lll
        rr
        rr
        rr
        rr
        rr
        rr
    }
        \toprule
        &&
            & \multicolumn{2}{c}{\tennis}
            & \multicolumn{2}{c}{\fscomp}
            & \multicolumn{2}{c}{\fsperf}
            & \multicolumn{2}{c}{\finediving}
            & \multicolumn{4}{c}{\finegym}
            \\
        &&
            & \multicolumn{2}{c}{}
            & \multicolumn{2}{c}{}
            & \multicolumn{2}{c}{}
            & \multicolumn{2}{c}{}
            & \multicolumn{2}{c}{Full}
            & \multicolumn{2}{c}{Start}
            \\
        &&
            & \multicolumn{1}{c}{$\delta$=1}
            & \multicolumn{1}{c}{2}
            & \multicolumn{1}{c}{1}
            & \multicolumn{1}{c}{2}
            & \multicolumn{1}{c}{1}
            & \multicolumn{1}{c}{2}
            & \multicolumn{1}{c}{1}
            & \multicolumn{1}{c}{2}
            & \multicolumn{1}{c}{1}
            & \multicolumn{1}{c}{2}
            & \multicolumn{1}{c}{1}
            & \multicolumn{1}{c}{2}
            \\
        \midrule

        \multicolumn{3}{l}{{\bf Default:} \OURMETHOD 200MF (RGB)}
            & 96.1 & \nms 97.7
            & \nms 81.0 & \nms 93.5 
            & \nms \sota{85.1} & \nms 95.7 
            & \sota{68.4} & \nms \sota{85.3}
            & \nms 47.9 & \nms 65.2
            & \nms 61.0 & \nms 78.4
            \\
        \multicolumn{3}{l}{{\bf Best:} \OURMETHOD 800MF (2-stream)}
            & \nms \sota{96.9} & \nms \sota{98.1}
            & \nms \sota{83.4} & \nms \sota{94.9} 
            & \nms 83.3 & \nms \sota{96.0} 
            & \nms 66.4 & \nms 84.8
            & \nms \sota{51.8} & \nms \sota{68.5}
            & \nms \sota{65.3} & \nms \sota{81.6}
            \\

        \midrule
        \multicolumn{1}{c}{Feature} & \multicolumn{1}{c}{Model} &
            \multicolumn{4}{l}{Extra loss (if any)} \\

        \midrule
        I3D~\cite{i3d}
        & MS-TCN &
            & 62.7 & 75.0 & 60.8 & \nms 79.1 & 64.0 & \nms 83.6 & - & - & - & - & - & - \\

        (RGB + flow)
        & MS-TCN & CALF
            & 59.7 & 73.6 & 56.4 & \nms 72.2 & 61.6 & \nms 81.5 & - & - & - & - & - & - \\

        & MS-TCN & dilate 1
            & 58.1 & \nms 75.4 & 59.7 & \nms 79.5 & \best{69.0} & \nms 89.3 & - & - & - & - & - & - \\

        \cmidrule{2-15}
        & GRU &
            & 40.7 & \nms 66.1 & 38.6 & \nms 58.7 & 41.4 & \nms 64.2 & - & - & - & - & - & - \\

        & GRU & CALF
            & \nms 45.7 & \nms 70.5 & \nms 31.2 & \nms 53.0 & \nms 50.5 & \nms 75.4 & - & - & - & - & - & - \\

        & GRU & dilate 1
            & \nms 41.5 & \nms 68.2 & 41.8 & \nms 69.8 & 52.5 & \nms 77.5 & - & - & - & - & - & - \\

        \cmidrule{2-15}
        & ASFormer &
            & 55.4 & \nms 74.5 & 60.8 & \nms 82.2 & \best{69.0} & \nms 88.8 & - & - & - & - & - & - \\
        & ASFormer & CALF
            & 58.1 & \nms 76.5 & \best{61.2} & \nms \best{82.4} & 66.6 & \nms \best{89.7} & - & - & - & - & - & - \\
        & ASFormer & dilate 1
            & 49.6 & \nms 72.9 & 58.1 & \nms 81.1 & 64.6 & \nms 87.5 & - & - & - & - & - & - \\

        \cmidrule{2-15}
        & TCN &
            & \nms 58.9 & \nms 75.1 & \nms 53.0 & \nms 72.0 & \nms 58.7 & \nms 81.3 & - & - & - & - & - & - \\

        & GCN &
            & \nms 42.6 & \nms 55.2 & \nms 19.9 & \nms 32.5 & \nms 27.1 & \nms 45.5 & - & - & - & - & - & - \\

        \cmidrule{2-15}
        & StridedTF &
            & \nms 34.3 & \nms 48.0 & \nms 27.0 & \nms 43.8 & \nms 40.5 & \nms 63.6 & - & - & - & - & - & - \\

        & StridedTF & dilate 1
            & \nms 44.8 & \nms 62.9 & \nms 36.2 & \nms 56.2 & \nms 47.2 & \nms 68.9 & - & - & - & - & - & - \\

        \cmidrule{2-15}
        MViT-B~\cite{mvit}
        & MS-TCN &
            & \best{67.0} & \nms 78.3 & 56.9 & \nms 75.8 & 63.6 & \nms 80.8 & 56.1 & \nms 73.9 & 31.0 & \nms 48.2 & \nms \best{41.7} & \nms 63.2 \\

        (RGB)
        & MS-TCN & CALF
            & 66.8 & \nms 79.3 & 57.4 & \nms 75.8 & 64.8 & \nms 84.3 & 56.3 & \nms 75.5 & 30.0 & \nms 48.3 & 40.1 & \nms 63.0 \\

        & MS-TCN & dilate 1
            & 64.0 & \nms 80.1 & 55.6 & \nms 79.9 & 62.1 & \nms 82.9 & \best{59.3} & \nms \best{78.3} & 28.7 & \nms 48.6 & \nms 40.5 & \nms \best{64.8} \\

        \cmidrule{2-15}
        & GRU &
            & 64.8 & 79.6 & 45.6 & \nms 69.6 & 56.8 & \nms 76.1 & 57.3 & 76.7 & \nms 25.9 & \nms 42.1 & \nms 34.0 & \nms 54.3 \\

        & GRU & CALF
            & 59.1 & \nms 76.4 & \nms 45.5 & \nms 71.1 & 52.9 & \nms 77.3 & 55.8 & 75.6 & \nms 20.1 & \nms 34.4 & \nms 27.0 & \nms 45.3 \\

        & GRU & dilate 1
            & \nms 61.4 & \nms \best{80.8} & 44.7 & \nms 73.1 & 55.1 & \nms 79.1 & 48.7 & \nms 76.5 & \nms 28.5 & \nms 48.6 & \nms 39.1 & \nms 62.2 \\

        \cmidrule{2-15}
        & ASFormer &
            & 63.2 & \nms 79.9 & 55.8 & \nms 81.5 & 54.9 & \nms 80.4 & 37.4 & \nms 67.1 & \nms 24.9 & \nms 42.5 & \nms 32.4 & \nms 52.9 \\
        & ASFormer & CALF
            & 63.9 & \nms 79.5 & 52.3 & \nms 76.6 & 55.7 & \nms 81.7 & 38.5 & \nms 67.4 & \nms 25.3 & \nms 42.9 & \nms 32.3 & \nms 53.8 \\
        & ASFormer & dilate 1
            & 58.0 & \nms 78.9 & \nms 53.9 & \nms 81.8 & 56.4 & \nms 79.9 & \nms 35.2 & \nms 65.5 & \nms 23.4 & \nms 42.1 & \nms 32.5 & \nms 55.3 \\

        \cmidrule{2-15}
        & TCN &
            & \nms 66.1 & \nms 80.4 & \nms 47.8 & \nms 67.9 & \nms 59.6 & \nms 80.2 & \nms 55.5 & \nms 77.2 & \nms \best{31.4} & \nms \best{49.1} & \nms 40.7 & \nms 62.8 \\

        & GCN &
            & \nms 36.4 & \nms 54.0 & \nms 20.8 & \nms 34.9 & \nms 27.7 & \nms 45.8 & \nms 38.8 & \nms 59.9 & \nms 12.3 & \nms 22.0 & \nms 16.8 & \nms 29.3 \\

        \cmidrule{2-15}
        & StridedTF &
            & \nms 37.9 & \nms 54.9 & \nms 27.3 & \nms 45.7 & \nms 8.7 & \nms 15.2 & \nms 38.3 & \nms 64.7 & \nms 15.8 & \nms 25.4 & \nms 22.0 & \nms 34.3 \\

        & StridedTF & dilate 1
            & \nms 54.8 & \nms 73.0 & \nms 32.0 & \nms 50.7 & \nms 39.7 & \nms 59.7 & \nms 42.1 & \nms 68.6 & \nms 20.6 & \nms 35.8 & \nms 26.4 & \nms 45.6 \\

        \bottomrule
    \end{tabularx}
    }
\end{table*}

\begin{table*}[p]
    \centering
    \caption{{\bf Spotting performance (mAP @ $\delta$ frames)} with features fine-tuned on RGB inputs.
    \textdagger~indicates NMS. The best baseline scores are \best{underlined}.
    }
    \label{tab:supp_full_fine_tuned}
    \vspace{0.5em}
    {
    \tiny
    \begin{tabularx}{\textwidth}{lll
        rr
        rr
        rr
        rr
        rr
        rr
    }
        \toprule
        &&
            & \multicolumn{2}{c}{\tennis}
            & \multicolumn{2}{c}{\fscomp}
            & \multicolumn{2}{c}{\fsperf}
            & \multicolumn{2}{c}{\finediving}
            & \multicolumn{4}{c}{\finegym}
            \\
        &&
            & \multicolumn{2}{c}{}
            & \multicolumn{2}{c}{}
            & \multicolumn{2}{c}{}
            & \multicolumn{2}{c}{}
            & \multicolumn{2}{c}{Full}
            & \multicolumn{2}{c}{Start}
            \\
        &&
            & \multicolumn{1}{c}{$\delta$=1}
            & \multicolumn{1}{c}{2}
            & \multicolumn{1}{c}{1}
            & \multicolumn{1}{c}{2}
            & \multicolumn{1}{c}{1}
            & \multicolumn{1}{c}{2}
            & \multicolumn{1}{c}{1}
            & \multicolumn{1}{c}{2}
            & \multicolumn{1}{c}{1}
            & \multicolumn{1}{c}{2}
            & \multicolumn{1}{c}{1}
            & \multicolumn{1}{c}{2}
            \\
        \midrule

        \multicolumn{3}{l}{{\bf Default:} \OURMETHOD 200MF (RGB)}
            & 96.1 & \nms 97.7
            & \nms 81.0 & \nms 93.5 
            & \nms \sota{85.1} & \nms 95.7 
            & \sota{68.4} & \nms \sota{85.3}
            & \nms 47.9 & \nms 65.2
            & \nms 61.0 & \nms 78.4
            \\
        \multicolumn{3}{l}{{\bf Best:} \OURMETHOD 800MF (2-stream)}
            & \nms \sota{96.9} & \nms \sota{98.1}
            & \nms \sota{83.4} & \nms \sota{94.9} 
            & \nms 83.3 & \nms \sota{96.0} 
            & \nms 66.4 & \nms 84.8
            & \nms \sota{51.8} & \nms \sota{68.5}
            & \nms \sota{65.3} & \nms \sota{81.6}
            \\

        \midrule
        \multicolumn{1}{c}{Feature} & \multicolumn{1}{c}{Model} &
            \multicolumn{4}{l}{Extra loss (if any)} \\
        \midrule
        TSP~\cite{tsp}
        & MS-TCN &
            & 90.1 & \nms 94.6 & 72.4 & \nms 87.4 & 74.3 & \nms 89.4 & 55.5 & \nms 76.0 & \nms 40.5 & \nms 58.5 & \nms 53.9 & \nms 73.4 \\

        & MS-TCN & CALF
            & 90.9 & \nms 95.0 & 72.1 & \nms 87.8 & 76.8 & 89.9 & 54.2 & \nms 73.8 & 36.9 & \nms 57.4 & 47.5 & \nms 71.4 \\

        & MS-TCN & dilate 1
            & \nms 87.5 & \nms 95.1 & 67.0 & \nms 85.5 & 76.6 & \nms 89.3 & 57.7 & \nms 75.9 & \nms 37.8 & \nms 57.3 & \nms 53.2 & \nms 73.5 \\

        \cmidrule{2-15}
        & GRU &
            & 89.5 & 95.1 & 66.6 & \nms 83.9 & 75.5 & \nms 89.4 & 55.5 & 76.5 & \nms 38.4 & \nms 57.2 & \nms 49.8 & \nms 70.5 \\

        & GRU & CALF
            & 88.6 & \nms 94.9 & 64.4 & \nms 83.0 & \nms 60.1 & \nms 84.3 & 57.0 & 78.2 & \nms 36.1 & \nms 57.2 & \nms 44.3 & \nms 70.0 \\

        & GRU & dilate 1
            & \nms 89.3 & \nms 96.0 & \nms 68.4 & \nms 88.3 & \nms 69.6 & \nms 90.6 & \nms 53.2 & \nms 77.4 & \nms 38.7 & \nms 58.8 & \nms 53.2 & \nms \best{74.2} \\

        \cmidrule{2-15}
        & ASFormer &
            & 89.8 & \nms 94.8 & \best{77.7} & \nms \best{94.1} & \best{80.2} & \nms \best{94.5} & 47.1 & \nms 73.2 & \nms 38.8 & \nms 57.6 & \nms 51.1 & \nms 72.0 \\
        & ASFormer & CALF
            & 89.0 & \nms 95.5 & 73.4 & \nms 92.5 & 78.0 & \nms 94.2 & 51.3 & \nms 77.4 & \nms 38.6 & \nms 57.6 & \nms 50.3 & \nms 71.6 \\
        & ASFormer & dilate 1
            & \nms 86.9 & \nms 95.4 & \nms 72.2 & \nms 94.0 & 78.0 & \nms 94.0 & \nms 49.2 & \nms 76.4 & \nms 36.5 & \nms 57.6 & \nms 50.4 & \nms 72.9 \\

        \cmidrule{2-15}
        & TCN &
            & \nms 88.1 & \nms 94.5 & \nms 62.6 & \nms 79.0 & \nms 67.3 & \nms 86.2 & \nms 51.9 & \nms 75.7 & \nms 41.1 & \nms \best{59.6} & \nms 53.5 & \nms 73.7 \\

        & GCN &
            & \nms 85.7 & \nms 93.4 & \nms 52.9 & \nms 70.6 & \nms 53.5 & \nms 74.8 & \nms 48.9 & \nms 71.0 & \nms 33.2 & \nms 49.5 & \nms 43.3 & \nms 62.2 \\

        & NetVLAD++ &
            & \nms 55.5 & \nms 72.7 & - & - & - & - & - & - & - & - & - & - \\

        \cmidrule{2-15}
        & StridedTF &
            & \nms 83.0 & \nms 93.3 & \nms 53.8 & \nms 73.3 & \nms 55.3 & \nms 76.9 & \nms 46.7 & \nms 74.2 & \nms 31.5 & \nms 47.8 & \nms 42.6 & \nms 60.9 \\

        & StridedTF & dilate 1
            & \nms 86.0 & \nms 94.7 & \nms 61.2 & \nms 83.1 & \nms 65.3 & \nms 84.6 & \nms 46.6 & \nms 76.2 & \nms 31.7 & \nms 51.6 & \nms 39.6 & \nms 63.2 \\

        \cmidrule{2-15}
        (\NumClasses + 1)-VC
        & MS-TCN &
            & 91.1 & \nms 94.8 & 66.5 & \nms 77.2 & 73.6 & \nms 83.8 & \best{63.2} & \nms 81.4 & \nms 40.9 & \nms 57.9 & \nms 53.2 & \nms 71.9 \\

        & MS-TCN & CALF
            & 91.0 & \nms 94.5 & 60.8 & \nms 73.1 & 75.2 & \nms 86.7 & 59.0 & \nms 76.4 & \nms 38.6 & \nms 56.8 & \nms 50.1 & \nms 70.8 \\

        & MS-TCN & dilate 1
            & \nms 90.3 & \nms 95.1 & 60.3 & \nms 73.6 & 77.2 & \nms 89.9 & 60.4 & \nms \best{83.5} & \nms 39.2 & \nms 58.2 & \nms 53.1 & \nms 73.8 \\

        \cmidrule{2-15}
        & GRU &
            & \nms 90.8 & \nms 96.0 & \nms 61.1 & \nms 75.5 & 73.0 & \nms 86.5 & 60.0 & \nms 80.6 & \nms 41.1 & \nms 57.9 & \nms 54.3 & \nms 72.3 \\

        & GRU & CALF
            & \nms 88.2 & \nms 95.4 & \nms 62.4 & \nms 77.2 & \nms 73.3 & \nms 85.0 & 61.8 & \nms 80.5 & \nms 39.6 & \nms 55.3 & \nms 51.8 & \nms 69.5 \\

        & GRU & dilate 1
            & \nms 91.5 & \nms \best{96.2} & \nms 61.7 & \nms 78.9 & \nms 76.8 & \nms 89.4 & \nms 58.2 & \nms 82.6 & \nms 38.6 & \nms 57.5 & \nms 53.6 & \nms 73.6 \\

        \cmidrule{2-15}
        & ASFormer &
            & \best{92.1} & \nms 95.5 & 67.2 & \nms 79.0 & 77.1 & \nms 88.9 & \nms 56.9 & \nms 83.0 & \nms 40.0 & \nms 56.8 & \nms 52.4 & \nms 70.3 \\
        & ASFormer & CALF
            & 90.8 & \nms 94.5 & 67.6 & \nms 79.5 & 75.2 & \nms 88.3 & 58.9 & \nms 82.2 & \nms 40.0 & \nms 56.9 & \nms 52.9 & \nms 71.2 \\
        & ASFormer & dilate 1
            & \nms 91.6 & \nms \best{96.2} & \nms 65.5 & \nms 79.8 & 75.4 & \nms 89.8 & 58.8 & \nms \best{83.5} & \nms 38.1 & \nms 56.9 & \nms 53.6 & \nms 72.9 \\

        \cmidrule{2-15}
        & TCN &
            & \nms 91.9 & \nms 96.1 & \nms 58.8 & \nms 74.2 & \nms 74.5 & \nms 86.8 & \nms 58.6 & \nms 77.9 & \nms \best{42.0} & \nms 58.9 & \nms \best{54.6} & \nms 73.3 \\

        & GCN &
            & \nms 88.4 & \nms 94.2 & \nms 54.8 & \nms 68.0 & 72.6 & \nms 84.1 & \nms 55.3 & \nms 75.4 & \nms 32.6 & \nms 46.0 & \nms 43.2 & \nms 58.6 \\

        & NetVLAD++ &
            & \nms 18.2 & \nms 26.3 & - & - & - & - & - & - & - & -  & - & - \\

        \cmidrule{2-15}
        & StridedTF &
            & \nms 88.4 & \nms 94.2 & \nms 39.2 & \nms 61.2 & \nms 59.1 & \nms 78.2 & \nms 50.4 & \nms 75.6 & \nms 24.7 & \nms 36.5 & \nms 34.4 & \nms 48.2 \\

        & StridedTF & dilate 1
            & \nms 88.6 & \nms 95.2 & \nms 59.3 & \nms 77.2 & \nms 71.7 & \nms 87.6 & \nms 45.5 & \nms 75.3 & \nms 24.3 & \nms 39.2 & \nms 30.4 & \nms 48.3 \\

        \bottomrule
    \end{tabularx}
    }
\end{table*}

\renewcommand{\tabcolsep}{0.2cm}
\begin{table}[tp]
    \centering
    \caption{{\bf Spotting performance (mAP @ $\delta$ frames) on \fscomp and \fsperf} using pose features~\cite{vpd}, fine-tuned on RGB and optical flow.
    \textdagger~indicates NMS.
    SOTA results with pose features are \sota{bold}.}
    \label{tab:supp_full_result_pose}
    \vspace{0.5em}
    {
    \scriptsize
    \begin{tabularx}{0.87\textwidth}{lll
        rr
        rr
    }
        \toprule
        &&
            & \multicolumn{2}{c}{\fscomp}
            & \multicolumn{2}{c}{\fsperf}
            \\
        &&
            & \multicolumn{1}{c}{$\delta$=1}
            & \multicolumn{1}{c}{2}
            & \multicolumn{1}{c}{1}
            & \multicolumn{1}{c}{2}
            \\
        \midrule

        \multicolumn{3}{l}{{\bf Default:} \OURMETHOD 200MF (RGB)}
            & \nms 81.0 & \nms 93.5 
            & \nms \best{85.1} & \nms 95.7 
            \\
        \multicolumn{3}{l}{{\bf Best:} \OURMETHOD 800MF (2-stream)}
            & \nms \best{83.4} & \nms \best{94.9} 
            & \nms 83.3 & \nms \best{96.0} 
            \\

        \midrule
        \multicolumn{1}{c}{Feature}
            & \multicolumn{1}{c}{Model}
            & \multicolumn{1}{c}{Extra loss (if any)}
            \\

        \midrule
        2D-VPD~\cite{vpd}
        & MS-TCN &
            & 77.2 & \nms 90.8 & 83.1 & \nms 94.5 \\

        & MS-TCN & CALF
            & \sota{83.5} & \nms \sota{96.2} & \sota{85.2} & \nms 96.0 \\

        & MS-TCN & dilate 1
            & 81.7 & \nms 95.5 & 82.4 & \nms \sota{96.4} \\

        \cmidrule{2-7}
        & GRU &
            & \nms 74.4 & \nms 94.2 & \nms 77.4 & \nms 94.9 \\

        & GRU & CALF
            & \nms 72.2 & \nms 93.4 & \nms 46.3 & \nms 63.0 \\

        & GRU & dilate 1
            & \nms 75.9 & \nms 94.3 & \nms 75.7 & \nms 94.1 \\

        \cmidrule{2-7}
        & ASFormer &
            & 78.8 & \nms 94.8 & 76.9 & \nms 95.1 \\
        & ASFormer & CALF
            & 78.2 & \nms 94.5 & 77.2 & \nms 93.9 \\
        & ASFormer & dilate 1
            & \nms 79.0 & \nms 95.7 & 79.3 & \nms 93.2 \\

        \cmidrule{2-7}
        & TCN &
            & \nms 75.0 & \nms 89.5 & \nms 76.5 & \nms 89.7 \\

        & GCN &
            & \nms 60.3 & \nms 72.5 & \nms 64.1 & \nms 77.2 \\

        \cmidrule{2-7}
        & StridedTF &
            & \nms 12.7 & \nms 20.0 & \nms 26.0 & \nms 37.1 \\

        & StridedTF & dilate 1
            & \nms 61.3 & \nms 79.2 & \nms 66.6 & \nms 84.2 \\

        \cmidrule{2-7}
        VI-VPD~\cite{vpd}
        & MS-TCN &
            & 73.4 & \nms 88.8 & 80.8 & \nms 91.9 \\

        & MS-TCN & CALF
            & 74.3 & 88.2 & 79.4 & \nms 91.3 \\

        & MS-TCN & dilate 1
            & 77.8 & \nms 91.3 & 77.9 & \nms 92.7 \\

        \cmidrule{2-7}
        & GRU &
            & 76.0 & \nms 94.8 & 78.2 & \nms 94.2 \\

        & GRU & CALF
            & \nms 74.6 & \nms 93.7 & \nms 77.6 & \nms 93.5 \\

        & GRU & dilate 1
            & \nms 74.9 & \nms 93.9 & \nms 77.6 & \nms 95.3 \\

        \cmidrule{2-7}
        & ASFormer &
            & 77.4 & \nms 94.8 & \sota{85.2} & \nms 95.6 \\
        & ASFormer & CALF
            & 80.2 & \nms 94.5 & 84.2 & \nms 95.9 \\
        & ASFormer & dilate 1
            & 79.7 & \nms 95.1 & 80.9 & \nms 93.7 \\

        \cmidrule{2-7}
        & TCN &
            & \nms 68.3 & \nms 85.2 & \nms 73.9 & \nms 87.9 \\

        & GCN &
            & \nms 57.5 & \nms 71.6 & \nms 60.3 & \nms 71.6 \\


        \cmidrule{2-7}
        & StridedTF &
            & \nms 23.4 & \nms 35.0 & \nms 67.5 & \nms 82.0 \\

        & StridedTF & dilate 1
            & \nms 65.7 & \nms 82.3 & \nms 69.7 & \nms 87.7 \\

        \bottomrule
    \end{tabularx}
    }
\end{table}

%% file: src/supp/tab_nms_result.tex
\renewcommand{\tabcolsep}{0.17cm}
\begin{table*}[p]
    \centering
    \caption{{\bf Ablation of non-maximum suppression (NMS)} at different tolerances \Tolerance for various model and feature configurations.
    Best results per configuration are \best{underlined}.
    A spotting method's sensitivity to NMS can depend on the model (single vs. 2-stream), dataset, and feature type.
    The differences between NMS windows of 1 to 4 are also subtle, and a NMS window of 1 frame or none at all is often sufficient.
    }
    \label{tab:supp_nms_result}
    {
    \scriptsize
    \begin{tabularx}{\textwidth}{l
        rr
        rr
        rr
        rr
        rr
        rr
    }
        \toprule
            & \multicolumn{2}{c}{\tennis}
            & \multicolumn{2}{c}{\fscomp}
            & \multicolumn{2}{c}{\fsperf}
            & \multicolumn{2}{c}{\finediving}
            & \multicolumn{4}{c}{\finegym}
            \\
            & \multicolumn{2}{c}{}
            & \multicolumn{2}{c}{}
            & \multicolumn{2}{c}{}
            & \multicolumn{2}{c}{}
            & \multicolumn{2}{c}{Full}
            & \multicolumn{2}{c}{Start}
            \\
            & \multicolumn{1}{c}{$\delta$=1}
            & \multicolumn{1}{c}{2}
            & \multicolumn{1}{c}{1}
            & \multicolumn{1}{c}{2}
            & \multicolumn{1}{c}{1}
            & \multicolumn{1}{c}{2}
            & \multicolumn{1}{c}{1}
            & \multicolumn{1}{c}{2}
            & \multicolumn{1}{c}{1}
            & \multicolumn{1}{c}{2}
            & \multicolumn{1}{c}{1}
            & \multicolumn{1}{c}{2}
            \\
        \midrule
        \multicolumn{11}{l}{{\bf Default:} \OURMETHOD 200MF (RGB)} \\
            \hspace{0.06in} No NMS
                & \best{96.1} & 96.8
                & 56.2 & 58.9
                & 62.6 & 65.4
                & \best{68.4} & 84.9
                & 40.6 & 45.4
                & 51.9 & 57.3\\
            \hspace{0.06in} window = 1
                & \best{96.1} & 96.7
                & 81.0 & 93.5
                & \best{85.1} & \best{95.7}
                & 66.3 & \best{85.3}
                & \best{47.9} & \best{65.2}
                & \best{61.0} & \best{78.4} \\
            \hspace{0.06in} window = 2
                & 95.9 & \best{97.6}
                & \best{81.3} & \best{93.9}
                & 84.2 & 95.2
                & 62.1 & 83.9
                & 47.4 & 64.8
                & 60.5 & 78.1 \\
            \hspace{0.06in} window = 4
                & 95.7 & 97.4
                & 81.2 & 93.8
                & 84.1 & 95.1
                & 59.2 & 81.6
                & 47.0 & 64.2
                & 60.2 & 77.6 \\

        \midrule
        \multicolumn{13}{l}{{\bf Best:} \OURMETHOD 800MF (2-stream)} \\
            \hspace{0.06in} No NMS
                & 93.6 & 94.2
                & 55.6 & 58.1
                & 57.3 & 60.4
                & 66.1 & 80.8
                & 43.2 & 48.1
                & 55.3 & 60.8 \\
            \hspace{0.06in} window = 1
                & \best{96.9} & \best{98.1}
                & \best{83.4} & \best{94.9}
                & \best{83.3} & \best{96.0}
                & \best{66.4} & \best{84.7}
                & \best{51.8} & \best{68.5}
                & \best{65.3} & \best{81.6} \\
            \hspace{0.06in} window = 2
                & 96.7 & 98.1
                & 82.8 & \best{94.9}
                & 83.0 & 95.8
                & 62.5 & 83.1
                & 51.2 & 68.0
                & 64.9 & 81.3 \\
            \hspace{0.06in} window = 4
                & 96.6 & 98.0
                & 82.8 & \best{94.9}
                & 83.0 & 95.8
                & 59.9 & 81.0
                & 50.7 & 67.2
                & 64.6 & 80.9 \\

        \midrule
        \multicolumn{13}{l}{{\bf Baseline:} MS-TCN w/ TSP features} \\
            \hspace{0.06in} No NMS
                & \best{90.1} & 94.4
                & \best{72.4} & 83.9
                & \best{74.3} & 89.2
                & \best{55.5} & 72.7
                & 40.0 & 47.6
                & 51.9 & 60.5 \\
            \hspace{0.06in} window = 1
                & 87.6 & \best{94.6}
                & 68.2 & \best{87.4}
                & 68.1 & \best{89.4}
                & 50.9 & \best{76.0}
                & \best{40.5} & \best{58.5}
                & 53.9 & 73.4 \\
            \hspace{0.06in} window = 2
                & 87.3 & 94.4
                & 68.2 & 87.3
                & 68.1 & \best{89.4}
                & 49.3 & 75.2
                & \best{40.5} & \best{58.5}
                & \best{54.1} & \best{73.6} \\
            \hspace{0.06in} window = 4
                & 87.0 & 94.0
                & 68.2 & 87.3
                & 68.1 & \best{89.4}
                & 47.7 & 73.4
                & 40.4 & 58.3
                & 54.0 & 73.4 \\

        \midrule
        \multicolumn{13}{l}{{\bf Baseline:} ASFormer w/ TSP features} \\
            \hspace{0.06in} No NMS
                & \best{92.1} & 94.0 & \best{67.2} & 75.3 & \best{77.1} & 85.9 & 56.8 & 69.5 & 33.8 & 39.1 & 42.9 & 48.5 \\
            \hspace{0.06in} window = 1
                & 91.8 & \best{95.5} & 66.1 & \best{79.0} & 74.5 & \best{88.9} & \best{56.9} & \best{83.0} & \best{40.0} & \best{56.8} & \best{52.4} & \best{70.3} \\
            \hspace{0.06in} window = 2
                & 91.5 & 95.4 & 66.1 & \best{79.0} & 74.5 & \best{88.9} & 55.9 & 82.3 & 39.9 & 56.7 & 52.3 & \best{70.3} \\
            \hspace{0.06in} window = 4
                & 91.4 & 95.2 & 66.1 & \best{79.0} & 74.5 & \best{88.9} & 55.1 & 81.0 & 39.7 & 56.5 & 52.2 & 70.2 \\
        \bottomrule
    \end{tabularx}
    }
\end{table*}

%% file: src/supp/tab_frame_exact.tex
\renewcommand{\tabcolsep}{0.12cm}
\begin{table*}[p]
    \centering
    \caption{{\bf Spotting performance (mAP @ $\Tolerance=0$)}, when predicting the exact frame of human annotation.
    SOTA is \sota{bold}. Best results per-category are otherwise \best{underlined}.
    As noted in~\autoref{sub:supp_result_exact}, the conclusions that can be drawn from this table are limited because of ambiguity in the frame-level annotations. }
    \label{tab:supp_result_exact}
    {
    \scriptsize
    \begin{tabularx}{\textwidth}{lrrrrrr}
        \toprule
            & \multicolumn{1}{c}{\tennis}
            & \multicolumn{1}{c}{\fscomp}
            & \multicolumn{1}{c}{\fsperf}
            & \multicolumn{1}{c}{\finediving}
            & \multicolumn{2}{c}{\finegym}
            \\
            &
            &
            &
            &
            & \multicolumn{1}{c}{Full}
            & \multicolumn{1}{c}{Start}
            \\
        \midrule
        {\bf Default:} \OURMETHOD 200MF (RGB)
            & \sota{71.6}
            & 36.7
            & \sota{40.5}
            & 30.1
            & 22.4 & 27.5 \\
        {\bf Best:} \OURMETHOD 800MF (2-stream)
            & 69.1
            & \best{37.6}
            & 38.6
            & \sota{30.2}
            & \sota{23.7} & \sota{29.2} \\
        \midrule
        MS-TCN w/ TSP features
            & 50.0
            & 33.3
            & 34.0
            & 23.2
            & \best{19.7} & \best{25.3}\\
        \hspace{4em} w/ (\NumClasses + 1)-VC features
            & 61.0
            & 31.9
            & 36.7
            & 26.7
            & 19.2 & 24.0 \\
        \multicolumn{2}{l}{\hspace{4em} w/ 2D-VPD features \& CALF}
            - & \sota{43.1} & \best{38.9} & - & - & - \\

        GRU w/ TSP features
            & 42.4
            & 30.6
            & 27.4
            & 15.4
            & 18.6 & 23.5 \\
        \hspace{2.2em} w/ (\NumClasses + 1)-VC features
            & 56.2
            & 28.3
            & 36.3
            & 18.7
            & 19.2 & 24.5 \\
        \hspace{2.2em} w/ 2D-VPD features \& CALF
            & - & 32.8 & 20.3 & - & - & - \\

        ASFormer w/ TSP features
            & 51.6
            & 36.6
            & 37.4
            & 22.5
            & 18.6 & 23.6 \\
        \hspace{4.4em} w/ (\NumClasses + 1)-VC features
            & \best{62.8}
            & 31.8
            & 36.6
            & \best{27.4}
            & 18.6 & 23.5 \\
        \hspace{4.4em} w/ 2D-VPD features \& CALF
            & - & 35.4 & 35.1 & - & - & - \\
        \bottomrule
    \end{tabularx}
    }
\end{table*}

%% file: src/supp/fig_pr_curve.tex
\begin{figure*}[p]
    \centering
    \begin{subfigure}{\textwidth}
        \centering
        \includegraphics[width=0.32\textwidth]{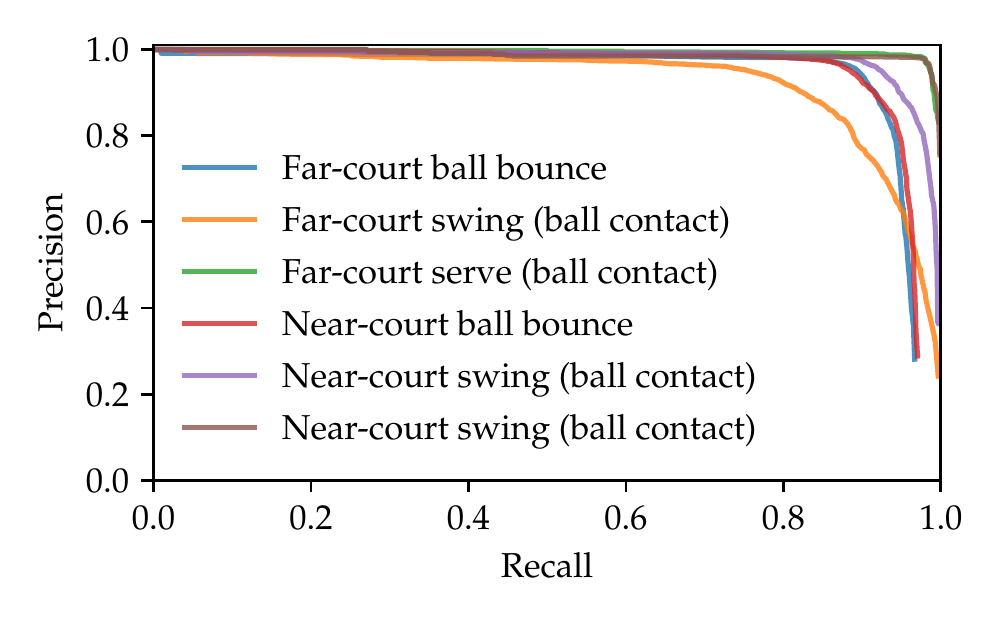}
        \includegraphics[width=0.32\textwidth]{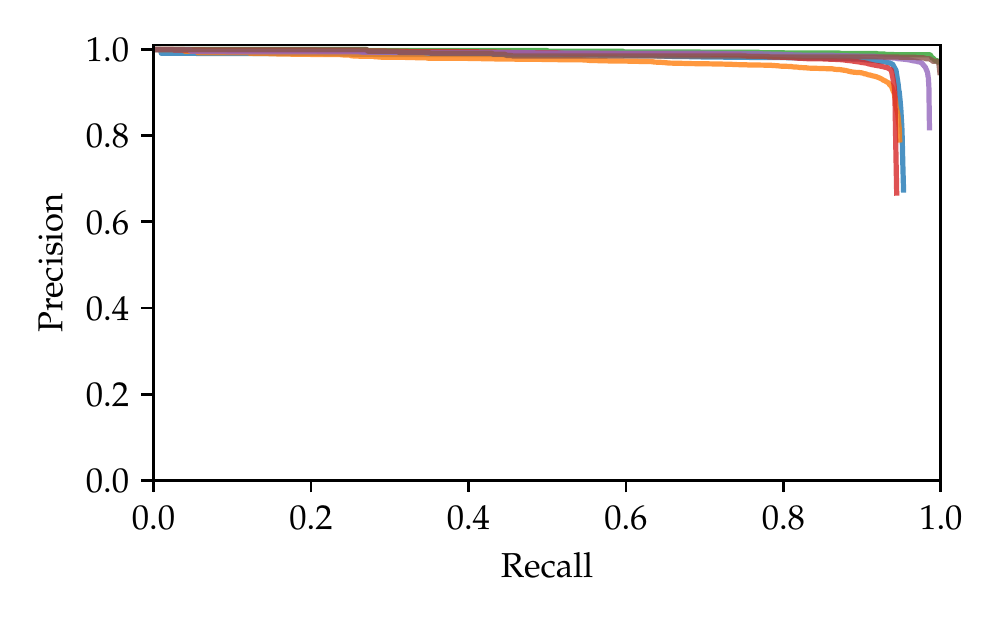}
        \vspace{-1em}
        \caption{Tennis (6 classes)}
    \end{subfigure}
    \begin{subfigure}{\textwidth}
        \centering
        \includegraphics[width=0.32\textwidth]{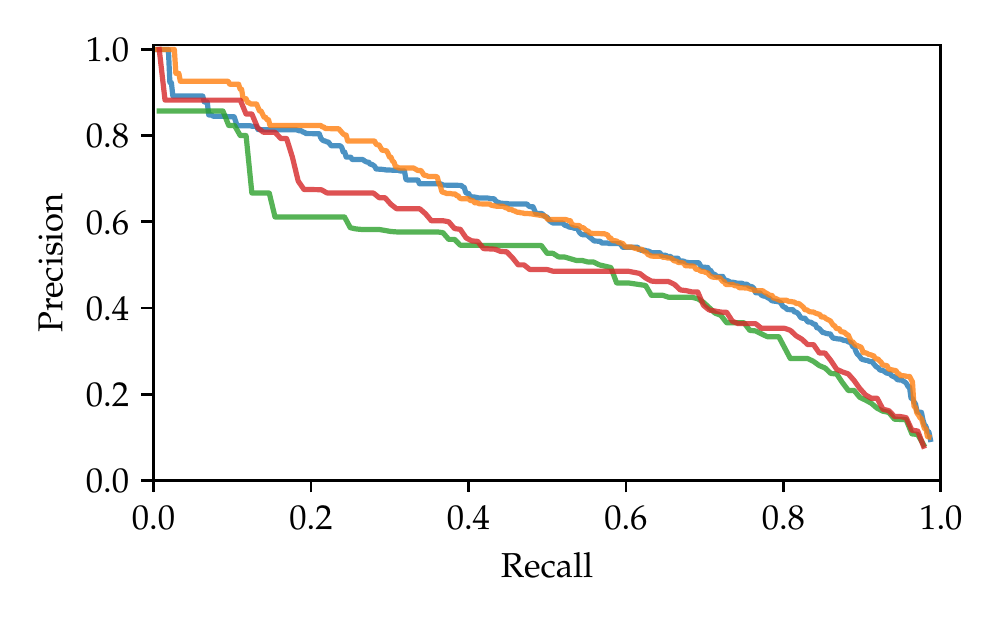}
        \includegraphics[width=0.32\textwidth]{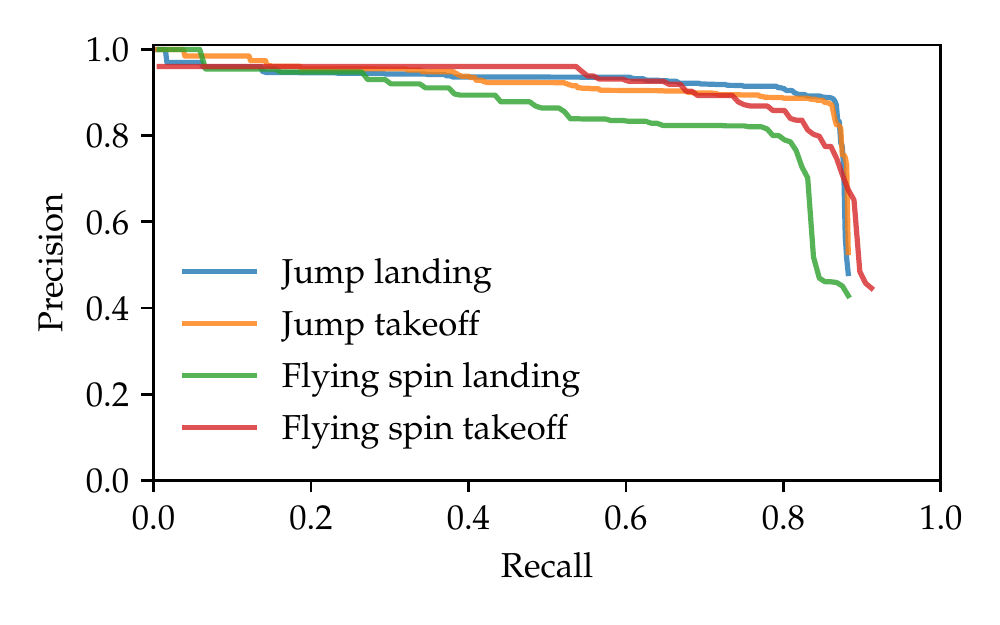}
        \vspace{-1em}
        \caption{\fscomp (4 classes)}
    \end{subfigure}
    \begin{subfigure}{\textwidth}
        \centering
        \includegraphics[width=0.32\textwidth]{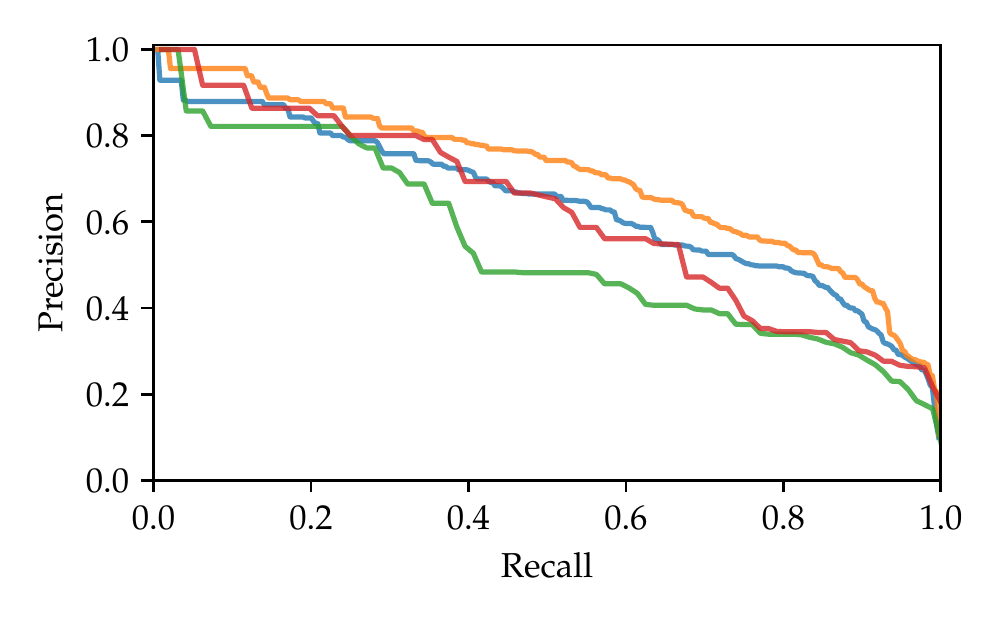}
        \includegraphics[width=0.32\textwidth]{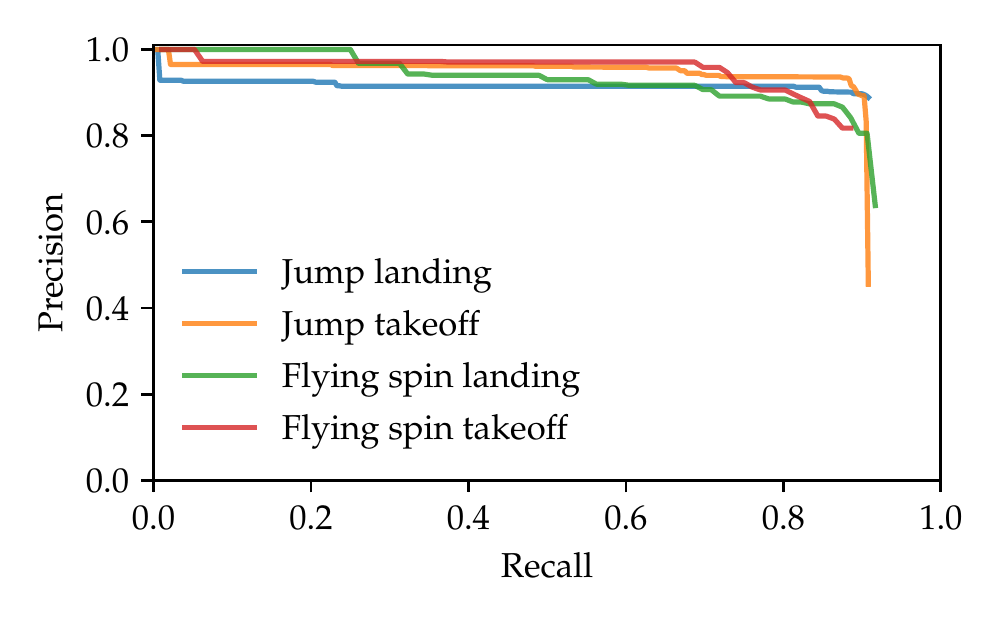}
        \vspace{-1em}
        \caption{\fsperf (4 classes)}
    \end{subfigure}
    \begin{subfigure}{\textwidth}
        \centering
        \includegraphics[width=0.32\textwidth]{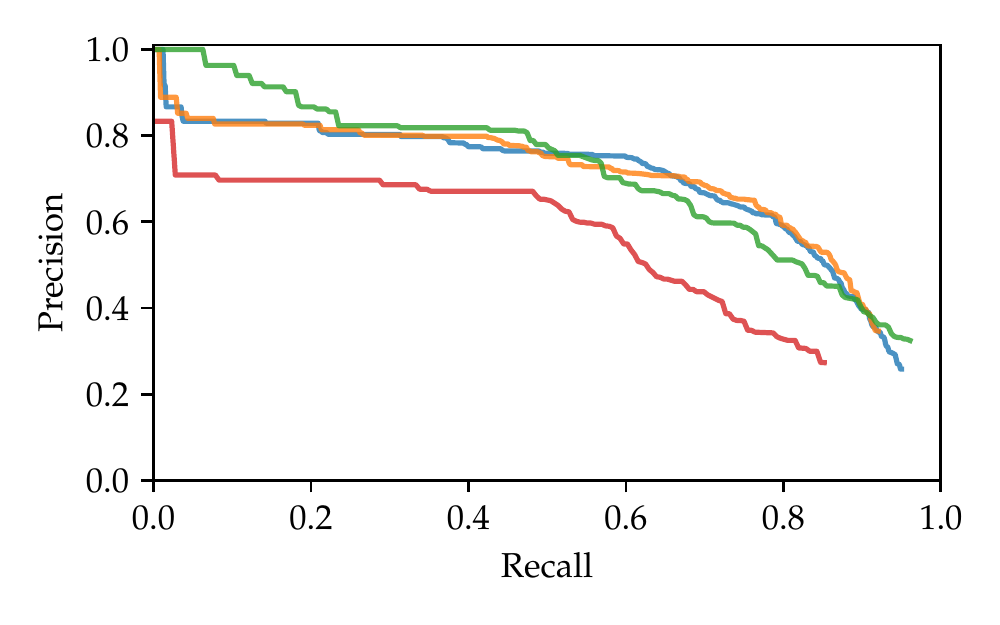}
        \includegraphics[width=0.32\textwidth]{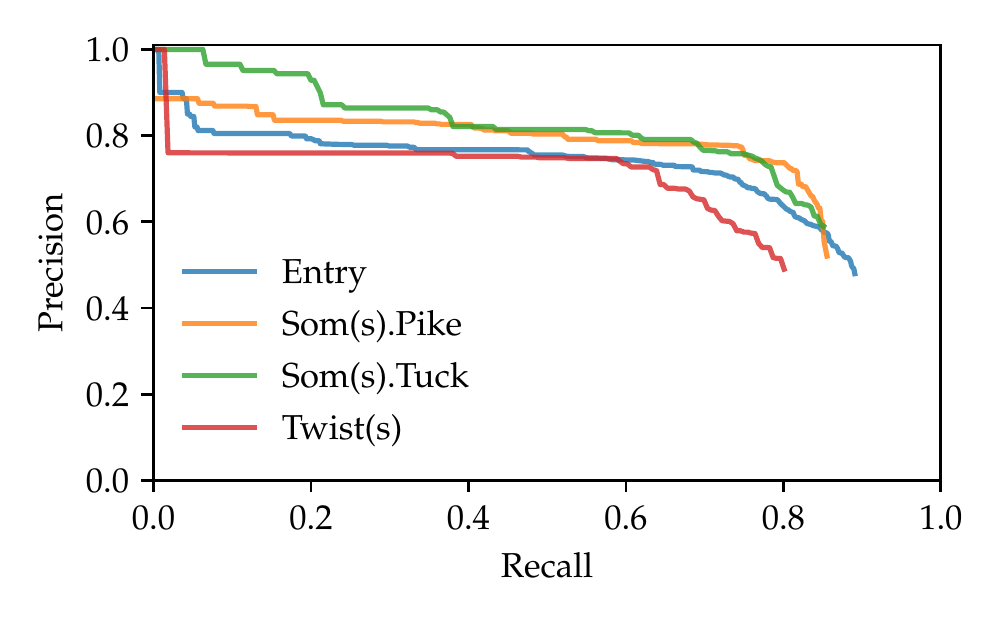}
        \vspace{-1em}
        \caption{\finediving (4 classes)}
    \end{subfigure}
    \begin{subfigure}{\textwidth}
        \centering
        \includegraphics[width=0.32\textwidth]{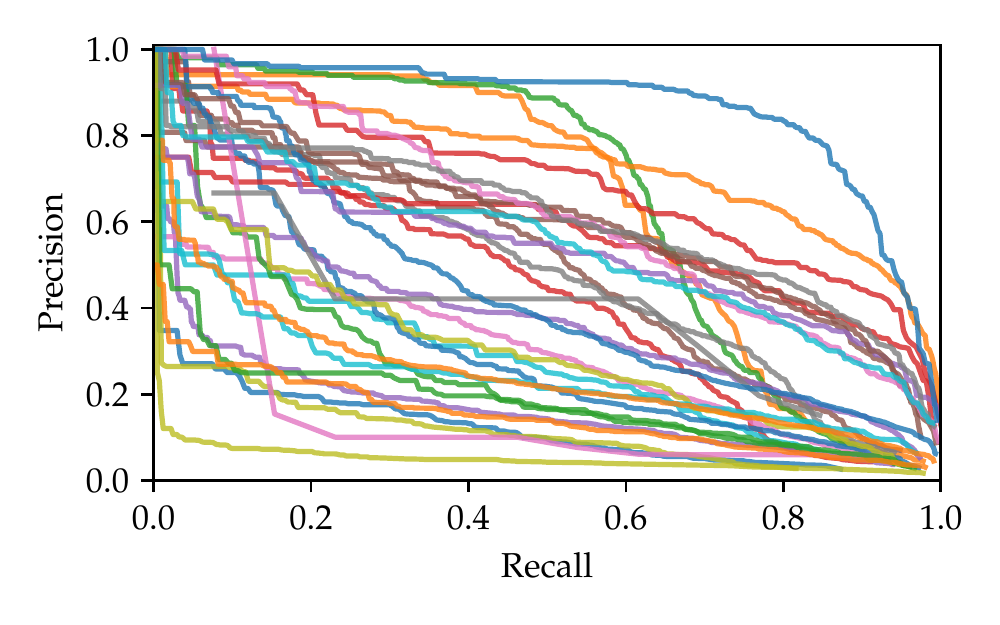}
        \includegraphics[width=0.32\textwidth]{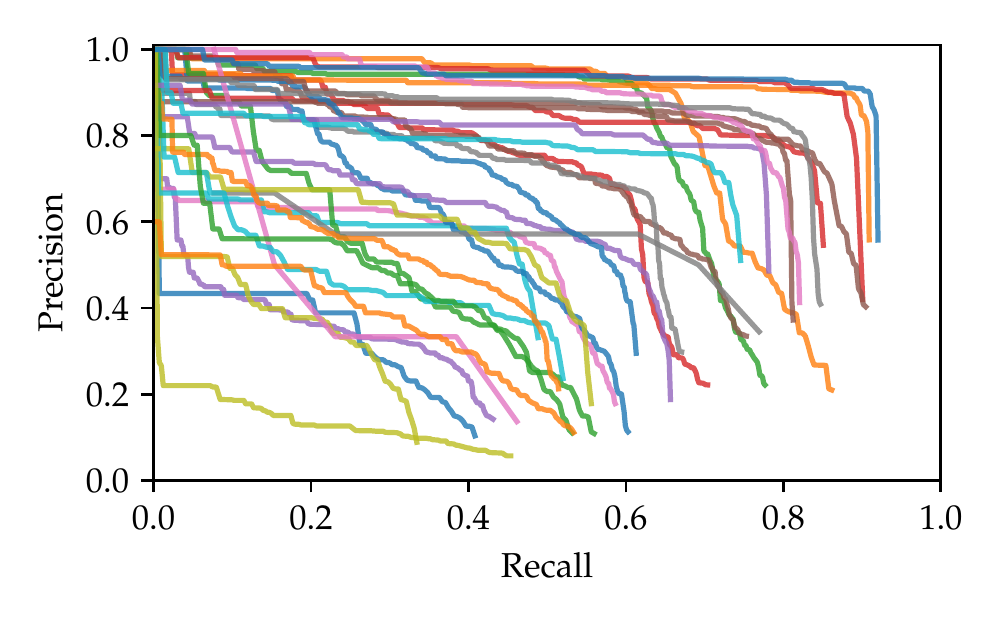}
        \vspace{-1em}
        \caption{\finegym-Full (32 classes)}
    \end{subfigure}
    \begin{subfigure}{\textwidth}
        \centering
        \includegraphics[width=0.32\textwidth]{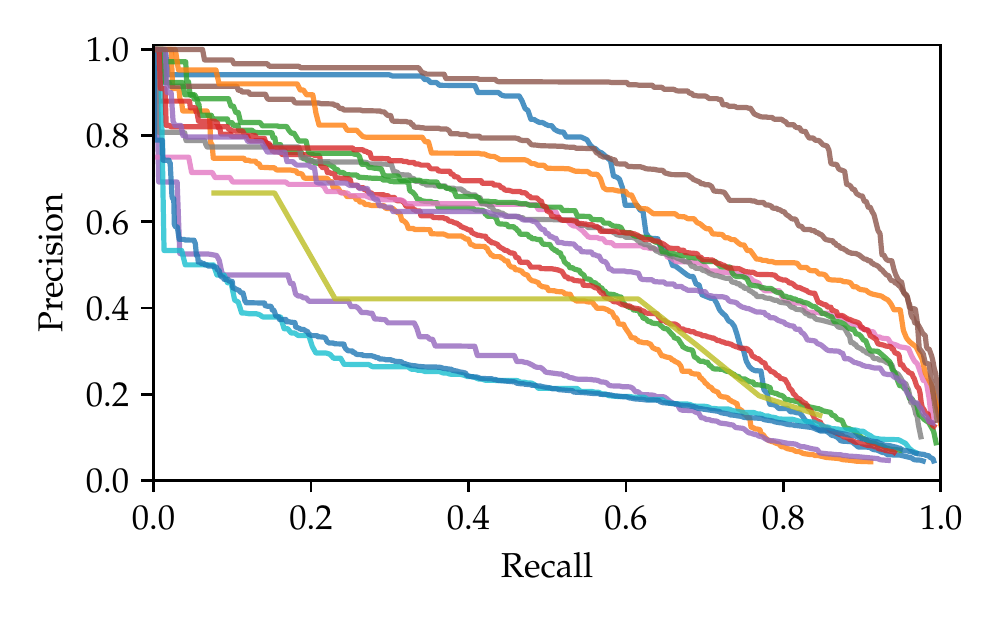}
        \includegraphics[width=0.32\textwidth]{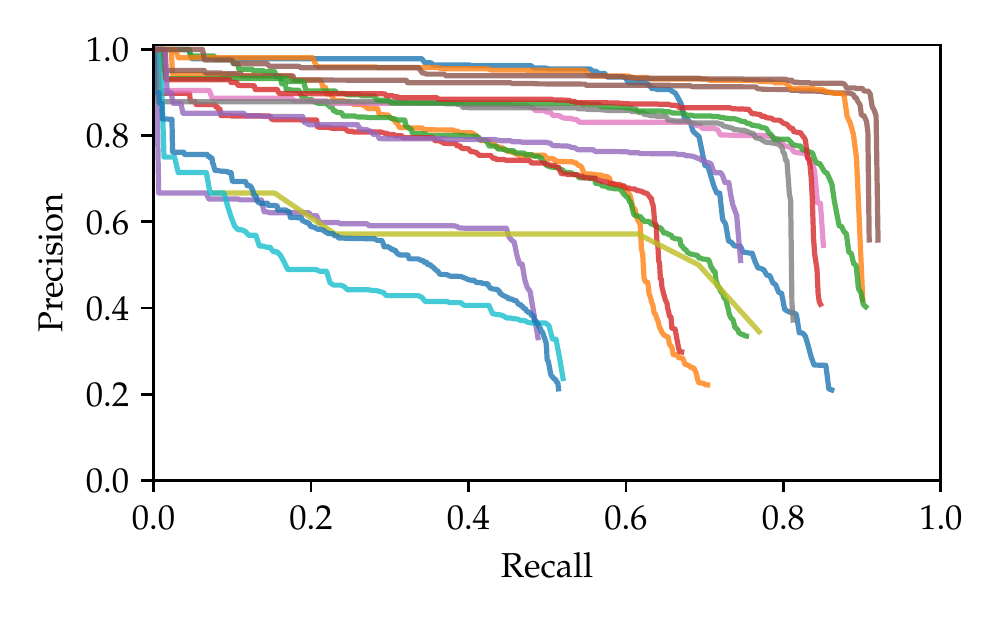}
        \vspace{-1em}
        \caption{\finegym-Start (16 classes)}
    \end{subfigure}

    \caption{
        {\bf Precision-recall curves for each event class at $\Tolerance = 1$}, produced by \OURMETHOD's default configuration. Charts on the {\bf left are without NMS} and charts on the {\bf right are with NMS}.
        NMS improves precision by suppressing nearby detections but can also lead to lower recall.
    }
    \label{fig:supp_pr_curves}
\end{figure*}

%% file: src/supp_dataset.tex
\section{Dataset Details}
\label{sec:supp_dataset}

We use the Tennis~\cite{vid2player}, Figure Skating~\cite{vpd}, \finediving~\cite{finediving}, and \finegym~\cite{finegym} datasets, with precise temporal event labels.

\subsection{Tennis}

The Tennis dataset is an extension of the dataset proposed by Zhang et al~\cite{vid2player} to 19 new videos.
Like the nine original videos from~\cite{vid2player}, the new videos are obtained from YouTube at full HD resolution and contain content from the US Open and Wimbledon tournaments.
We annotate at least one full `set' (a unit of gameplay) from each of the 19 new videos in order to diversify the dataset for training and evaluation.

To focus on the temporal aspect of precise spotting, we evaluate on the six top-level categories of events enumerated in~\autoref{sec:dataset} and~\autoref{tab:tennis_class}.
These events are selected by their temporal definitions instead of the full set of semantic action attributes (e.g., swing type differentiated by topspin vs. slice; forehand vs. backhand; volley).
The dataset contains 1.3M frames, of which 2.6\% are precise temporal events.

\subsection{Figure Skating}
\label{sub:supp_fs_dataset}

We extend the labels by Hong et al~\cite{vpd}, which include fine-grained action classes and their temporal extents at approximately 1 second precision.
To perform precise spotting, we manually re-annotate the labels to frame-accurate take-off and landing events.

As in Tennis, we separate temporally precise spotting from fine-grained classification of actions (e.g., the jump type) in order to focus on the temporal aspect of the spotting problem.
See~\autoref{tab:fs_class} for event statistics.
The dataset contains 1.6M frames, of which only 0.23\% are precise temporal events.

\subsection{FineDiving}

We use the pre-extracted frames provided by Xu et al~\cite{finediving} and spot the frames of transition between segments.
The events include somersaults.pike, somersaults.tuck, twists, and entry.
Note that we ignore the number of revolutions when generating frame-level event labels.
See~\autoref{tab:finediving_class} for event statistics.
The dataset contains 547K frames, of which 2.2\% are precise temporal events.

\subsection{FineGym}

\finegym is a large gymnastics video dataset released by Shao et al~\cite{finegym}.
It contains annotations for balance beam, floor exercises, uneven bars, and vaulting.
The dataset is primarily used for action recognition, with 288 fine-grained classes and their time intervals.
These actions are contained within individual performances (e.g., an untrimmed balance beam routine), and several performances appear in a single video from YouTube.

We detect precise events within the untrimmed performances and split the dataset three ways for training, validation, and testing; these splits do not contain overlap in performances and source videos.
We discard any performances that do not contain temporal annotations, have malformed annotations, or have annotations that are missing a class label in Gym288, leaving 5,374 performances.

Shao et al.~\cite{finegym} propose a hierarchy of action categories (to which the Gym288 classes belong), and we reduce the spotting problem to the granularity of these categories (e.g., ``balance beam dismounts'' is one example).
Because our focus is temporal precision, we leave the challenging task of (unbalanced) 288-way action classification, which can be performed after events have been spotted, to past and future work on fine-grained action recognition.

We define temporally precise events in \finegym as the start and end frames of action intervals.
This definition is straightforward for actions in balance beam, floor exercises, and uneven bars.
Each vault, however, is specified as a sequence of three back-to-back segments, which we convert into four events.
See~\autoref{tab:finegym_class} for the event breakdown and statistics.

A minority of videos (259) in the \finegym dataset have frame rates higher than 25--30 FPS.
For consistency, since our spotting tolerances are defined in \Tolerance frames, we resample those videos to between 25--30 FPS.
The final dataset contains 7.6M frames, 1.1\% of which are precise temporal events.

\input{src/supp/tab_dataset.tex}

%% file: src/supp/tab_dataset.tex
\begin{table}[p]
    \centering
    \caption{{\bf Tennis dataset:} event classes and their counts.}
    \label{tab:tennis_class}
    {
    \begin{tabularx}{0.605\columnwidth}{lrrr}
        \toprule
        \multicolumn{1}{c}{Event class} & Train & Val & Test \\
        \midrule
        Near-court serve (ball contact)
            & 673 & 238 & 779 \\
        Near-court swing (ball contact)
            & 2199 & 709 & 4136 \\
        Near-court ball bounce
            & 2606 & 871 & 4650 \\
        Far-court serve (ball contact)
            & 657 & 200 & 800 \\
        Far-court swing (ball contact)
            & 2220 & 757 & 4146 \\
        Far-court ball bounce
            & 2621 & 867 & 4662 \\
        \bottomrule
    \end{tabularx}
    }
\end{table}

\begin{table}[p]
    \centering
    \caption{{\bf Figure Skating dataset:} event classes and their counts.}
    \label{tab:fs_class}
    {
    \begin{tabularx}{0.68\columnwidth}{lrrrrrr}
        \toprule
            & \multicolumn{3}{c}{\fscomp} & \multicolumn{3}{c}{\fsperf} \\
        \multicolumn{1}{c}{Event class}
            & Train & Val & Test
            & Train & Val & Test \\
        \midrule
        Jump takeoff
            & 704 & 233 & 527
            & 723 & 372 & 369 \\
        Jump landing
            & 704 & 233 & 527
            & 723 & 372 & 369 \\
        Flying spin takeoff
            & 178 & 59 & 136
            & 183 & 94 & 96 \\
        Flying spin landing
            & 178 & 59 & 136
            & 183 & 94 & 96 \\
        \bottomrule
    \end{tabularx}
    }
\end{table}

\begin{table}[p]
    \centering
    \caption{{\bf \finediving dataset:} event classes and their counts.}
    \label{tab:finediving_class}
    {
    \begin{tabularx}{0.38\columnwidth}{lrrr}
        \toprule
        \multicolumn{1}{c}{Event class} & Train & Val & Test \\
        \midrule
        Entry
            & 1794 & 449 & 741 \\
        Som(s).Pike
            & 1254 & 345 & 553 \\
        Som(s).Tuck
            & 667 & 149 & 255 \\
        Twist(s)
            & 467 & 120 & 216 \\
        \bottomrule
    \end{tabularx}
    }
\end{table}

\def\checkmark{\tikz\fill[scale=0.25](0,.35) -- (.25,0) -- (1,.7) -- (.25,.15) -- cycle;}

\renewcommand{\tabcolsep}{0.225cm}
\begin{table*}[p]
    \centering
    \caption{{\bf FineGym dataset:} event classes and their counts.
    The classes are based on the `set-level categories' defined by Shao et al~\cite{finegym}.
    We refer to the full set of classes as \finegym-Full and a more visually consistent subset, containing primarily start events, as \finegym-Start.}
    \label{tab:finegym_class}
    {
    \begin{tabularx}{\columnwidth}{lcrrr}
        \toprule
        \multicolumn{1}{c}{Event class}
            & In \finegym-Start & Train & Val & Test \\
        \midrule
        Floor exercise leap\_jump\_hop start & \checkmark
            & 2007 & 602 & 629 \\
        Floor exercise leap\_jump\_hop end &
            & 2007 & 602 & 629 \\
        Floor exercise turns start & \checkmark
            & 683 & 197 & 223 \\
        Floor exercise turns end &
            & 683 & 197 & 223 \\
        Floor exercise side\_salto start & \checkmark
            & 23 & 13 & 13 \\
        Floor exercise side\_salto end &
            & 23 & 13 & 13 \\
        Floor exercise front\_salto start & \checkmark
            & 818 & 259 & 268 \\
        Floor exercise front\_salto end &
            & 818 & 259 & 268 \\
        Floor exercise back\_salto start & \checkmark
            & 1850 & 524 & 604 \\
        Floor exercise back\_salto end &
            & 1850 & 524 & 604 \\
        Balance beam leap\_jump\_hop start & \checkmark
            & 3062 & 765 & 960 \\
        Balance beam leap\_jump\_hop end &
            & 3062 & 765 & 960 \\
        Balance beam turns start & \checkmark
            & 857 & 215 & 299 \\
        Balance beam turns end &
            & 857 & 215 & 299 \\
        Balance beam flight\_salto start & \checkmark
            & 2637 & 720 & 830 \\
        Balance beam flight\_salto end &
            & 2637 & 720 & 830 \\
        Balance beam flight\_handspring start & \checkmark
            & 1835 & 440 & 618 \\
        Balance beam flight\_handspring end &
            & 1835 & 440 & 618 \\
        Balance beam dismounts start & \checkmark
            & 763 & 188 & 267 \\
        Balance beam dismounts end &
            & 763 & 188 & 267 \\
        Uneven bars circles start & \checkmark
            & 4143 & 1151 & 1318 \\
        Uneven bars circles end &
            & 4143 & 1151 & 1318 \\
        Uneven bars flight\_same\_bar start & \checkmark
            & 1029 & 270 & 325 \\
        Uneven bars flight\_same\_bar end &
            & 1029 & 270 & 325 \\
        Uneven bars transition\_flight start & \checkmark
            & 2079 & 630 & 680 \\
        Uneven bars transition\_flight end &
            & 2079 & 630 & 680 \\
        Uneven bars dismounts start & \checkmark
            & 750 & 225 & 252 \\
        Uneven bars dismounts end &
            & 750 & 225 & 252\\
        Vault (timestamp 0) &
            & 1263 & 367 & 401 \\
        Vault (timestamp 1) & \checkmark
            & 1263 & 367 & 401 \\
        Vault (timestamp 2) & \checkmark
            & 1263 & 367 & 401 \\
        Vault (timestamp 3) &
            & 1263 & 367 & 401 \\
        \bottomrule
    \end{tabularx}
    }
\end{table*}